\documentclass[accepted]{uai2026} 
\usepackage[american]{babel}
\usepackage{natbib}
    \renewcommand{\bibsection}{\subsubsection*{References}}
\usepackage{mathtools} 
\usepackage{booktabs} 
\usepackage{tikz} 
\usepackage{amssymb}

\usepackage{amsmath,amsfonts,bm}
\usepackage[ruled,vlined,linesnumbered]{algorithm2e}
\usepackage{float}
\usepackage{subcaption}
\usepackage{multirow}
\usepackage{pifont}              
\usepackage{amsthm}
\usepackage{bbm}
\usepackage{fontawesome5} 

\usepackage{blindtext}

\usepackage{graphicx}

\definecolor{ForestGreen}{RGB}{34,139,34}    

\newcommand{\CC}{\mathcal{C}}
\newcommand{\DC}{\mathcal{D}}

\newcommand{\NC}{\mathcal{N}}
\newcommand{\SC}{\mathcal{S}}
\newcommand{\XC}{\mathcal{X}}
\newcommand{\YC}{\mathcal{Y}}
\newcommand{\TC}{\mathcal{T}}

\newcommand{\RR}{\mathbb{R}}
\newcommand{\EE}{\mathbb{E}}

\newcommand{\pmodel}{\widehat{p}}

\newcommand{\ptrue}{p^\star}

\newcommand{\HPRa}{\text{HPR}_{1-\alpha}}
\def\ceil#1{\lceil #1 \rceil}
\newcommand{\indicator}{\mathbbm{1}}
\newcommand{\softmax}{\text{Softmax}}

\def\Pr{P}

\newtheorem{definition}{Definition}

\newcommand{\spm}{\ensuremath{\mathbin{\scriptstyle\pm}}} 

\DeclareMathOperator*{\argmax}{arg\,max}

\newtheorem{theorem}{Theorem}[section]

\newtheorem{lemma}[theorem]{Lemma}
\newtheorem{remark}[theorem]{Remark}

\definecolor{ForestGreen}{HTML}{008B8B}
\definecolor{red}{HTML}{D95F02}

\title{Adaptive Cumulative Mass Calibration with Conformal Prediction}

\author[1]{\href{mailto:<daniil.kazantsev@mbzuai.ac.ae>?Subject=Your UAI 2026 paper}{Daniil~Kazantsev}{}}
\author[1,2]{Eric~Moulines}
\author[1]{\href{mailto:<maxim.panov@mbzuai.ac.ae>?Subject=Your UAI 2026 paper}{Maxim~Panov}}
\author[1]{\href{mailto:<nikita.kotelevskii@mbzuai.ac.ae>?Subject=Your UAI 2026 paper}{Nikita~Kotelevskii}}
\author[1]{Mohsen~Guizani}
\affil[1]{%
   Mohamed bin Zayed University of Artificial Intelligence (MBZUAI)\\
   Abu Dhabi, United Arab Emirates
}
\affil[2]{%
    École pour l'informatique et les techniques avancées (EPITA), Paris, France
}
  
\begin{document}
\maketitle

\begin{abstract} 
  Reliable probability estimates by classifiers are essential in high-risk applications. 
  In practice, however, predicted probabilities are often miscalibrated, and many existing post-hoc calibration methods typically lack guarantees that a specific notion of calibration is achieved after the correction procedure is applied. 
  We introduce a \emph{set-based} perspective on calibration through the notion of \emph{cumulative mass calibration} and the corresponding error measures. 
  We propose a new calibration procedure based on conformal prediction that forms cumulative probabilities with guaranteed marginal coverage. 
  We introduce an \emph{adaptive} temperature scaling algorithm, with the temperature tuned for each input to satisfy the conformal coverage constraint. 
  As we show, this procedure can be efficiently implemented. 
  Across image classification tasks, particularly in settings with many classes, our method improves newly introduced calibration error measures (CMCE and \(\alpha\)-CMCE) \emph{and} standard metrics (such as ECE, cw-ECE, MCE) over the existing baselines.
  
  The code is publicly available on \href{https://github.com}{GitHub~\faGithub~\footnotemark}.
\end{abstract}

\footnotetext{\tiny{\url{https://github.com/stat-ml/conformal_probability_calibration}}}


\section{Introduction}
\label{sec:introduction}

  Classification models, particularly deep neural networks, are widely used in high-risk applications, including medical diagnosis and autonomous driving~\citep{bojarski2016end,esteva2017dermatologist}.
  In such settings, accurate probability estimates are crucial for decision-making.
  For example, when a model assigns a small probability to an outcome, a system might abstain from predicting or alert the user about the risk of error~\citep{geifman2017selective,hullermeier2021aleatoric,franc2023optimal}.
  However, modern neural networks are notoriously miscalibrated, and their predicted probabilities are often unreliable~\citep{guo2017calibration,kull2019beyond,zadrozny2002transforming}.

  The issue above has motivated extensive research on understanding and improving calibration. 
  A key difficulty is that, given a finite dataset of one-hot labels, achieving and measuring \emph{full multiclass (distributional) calibration} (see Definition~\ref{def:distribution-calibration}) is practically challenging.

  \begin{figure*}[!htb]
    \centering
    \adjustbox{width=\textwidth,center,trim=35mm 0mm 0mm 0mm,clip}{\input{pics/flow_chart}}
    \caption{Flowchart of the proposed approach. In the first stage (calibrator fitting), a conformal predictor is fitted. In the second stage (calibration), the fitted conformal predictor is first applied to obtain the conformal set \(C_\alpha(x)\). We then use binary search to determine \( \tau^{\star} \), an approximate solution to an equation whose right-hand side is the constant \(1-\alpha\) and whose left-hand side is the total predicted probability mass of the elements in \(C_\alpha(x)\), which is monotonically decreasing in \( \tau \).
    }
    \label{fig:flow_chart}
  \end{figure*}
  
  Therefore, in practice, more relaxed notions of calibration are used instead.
  For instance, \citet{guo2017calibration} showed that modern models tend to be overconfident and studied \emph{confidence} calibration, using expected calibration error (ECE) as the metric. \citet{kull2019beyond} introduced \emph{class-wise} calibration and the corresponding class-wise ECE (cw-ECE).
  Many other notions of calibration and associated metrics have been proposed~\citep{nixon2019measuring,widmann2019calibration,zhao2021calibrating}.

  However, existing post-hoc calibration methods suffer from two significant problems.
  First, they typically do not directly target the associated calibration metric and thus provide no guarantee that the metric will improve~\citep{vaicenavicius2019evaluating,gupta2020distribution}. 
  For example, temperature scaling~\citep{guo2017calibration} does not guaranty of the improvement of ECE. and Dirichlet calibration~\citep{kull2019beyond} does not provably reduce cw-ECE.
  Second, methods based on one-vs-rest reductions (see, e.g., \citep{zadrozny2002transforming}) require fitting a separate calibrator per class, which becomes data-intensive as the number of classes grows~\citep{le2024confidence}.
  Furthermore, existing notions of calibration have not been formulated in terms of cumulative sums of predicted probabilities.
  This is a consequential gap, as such cumulative quantities play an important and crucial role in decision-making~\citep{zhao2021calibrating,kiyani2025decision}.

  In this paper, we address these challenges and introduce new notions of calibration, corresponding error measures, and several practical post-hoc calibration procedures.
  Our key method is based on \emph{conformal prediction} (CP; \citealp{shafer2008tutorial,angelopoulos2021gentle}). Specifically, we introduce a method that uses CP sets to calibrate probabilities via input \textit{adaptive} temperature scaling. We show how to implement this procedure efficiently with one-dimensional binary search.
  Notably, the conformal-based calibration improves \(\alpha\)-cumulative mass calibration (see Definition~\ref{def:alpha_mass_calibration}), and provides distribution-free marginal coverage guarantees; see equation~\eqref{eq:cmce-def}.
  We also introduce an \(\alpha\)-agnostic notion of calibration and its corresponding error measure, the \emph{Cumulative Mass Calibration Error (CMCE)}, which considers calibration quality across all confidence levels.
  %

  Our main \textbf{contributions} are as follows. 
  \begin{enumerate}
    \item We introduce new notions of calibration, defined for \textit{cumulative masses}, that are necessary conditions for full multiclass (distributional) calibration; see Section~\ref{sec:mass_calibration}. 

    \item We develop an efficient post-hoc calibration method that, based on conformal prediction sets, transforms predicted probabilities via \emph{adaptive} temperature-scaling, improving \(\alpha\)-mass calibration (see Definition~\ref{def:alpha_mass_calibration}) and corresponding error measures; see equation~\eqref{eq:cmce-def} and Section~\ref{sec:method}.

    \item We propose a confidence level agnostic calibration measure, \emph{Cumulative Mass Calibration Error (CMCE)}, and a method that directly optimizes it; see Section~\ref{sec:cmce}.

    \item We demonstrate through large-scale experiments that our algorithms improve CMCE \emph{and} other classical calibration errors, despite not being designed specifically for them; see Section~\ref{sec:experiments}.
  \end{enumerate}


\section{Related Work}
\label{sec:related_work}
  Deep neural networks often provide \emph{miscalibrated} probabilities, which pose a challenge to their reliable use in practice.
 For example, predicted confidences may not match empirical accuracies~\citep{guo2017calibration}, class frequencies may be inaccurately estimated~\citep{kull2019beyond}, and output can be unreliable for decision-making~\citep{zhao2021calibrating}, among other issues. 
  These problems motivate different practical notions of calibration,  algorithms, and evaluation metrics.

  Two general strategies have been proposed for improving calibration: (i) training-time methods (e.g., label smoothing~\citep{muller2019does}), which require a specialized training procedure, and (ii) post-hoc calibration of an already pretrained model.
  While training-time methods can be effective, they modify the training process and often require retraining of the model from scratch. 
  Consequently, post-hoc calibration methods are often preferred because they work directly with pretrained models.

  The classical post-hoc calibration methods include Temperature Scaling~\citep{guo2017calibration}, Platt Scaling~\citep{platt1999probabilistic}, Isotonic Regression~\citep{zadrozny2002transforming}, and some parametric extensions such as Dirichlet Calibration~\citep{kull2019beyond} and Adaptive Temperature Scaling~\citep{joy2023sample}.
  These methods are simple, widely used, and improve specific notions of calibration.

  Some methods can be directly applied to a multiclass setting. For example, the classical temperature scaling fits a single global \(\tau\) for all classes and covariates simultaneously.
  In contrast, some other approaches may work for multiclass classification indirectly by first fitting one-vs-rest binary classifiers and then aggregating them into a multiclass classifier. 
  Therefore, when the number of classes \(K\) is large, these reductions become data-demanding~\citep{le2024confidence} and can be impractical.

  The classical temperature scaling, however, applies a single temperature to all inputs \(x\), resulting in a non-adaptive procedure.
  To mitigate this issue, adaptive methods were proposed that learn input- or class-dependent temperatures~\citep{tomani2022parameterized,joy2023sample,balanya2024adaptive}.  
  These methods increase expressiveness but introduce additional models and hyperparameters, risk overfitting small calibration sets, and typically lack formal calibration guarantees.

  Conformal prediction constructs \emph{prediction sets} with guaranteed \emph{marginal} coverage under the data exchangeability assumption~\citep{shafer2008tutorial,angelopoulos2021gentle,plassier2024probabilistic,plassier2024efficient}.
  While exact conditional guarantees are impossible without additional assumptions~\citep{vovk2012conditional,foygel2021limits}, recent work shows that \(x\)-dependent but invertible score transforms can improve conditional behavior while retaining marginal validity~\citep{colombo2024normalizing,plassierrectifying}.  
  However, standard CP returns sets rather than calibrated probability vectors. Recent studies examine the converse interaction, namely how calibration affects set efficiency~\citep{huang2024conformal,xi2024confidencecalibration}, but not how conformal sets can be leveraged to produce calibrated probabilities (the direction we study in this paper).


\section{Classifier Calibration}
\label{sec:background}
  We begin by introducing the necessary notations.
  We denote by \((X,Y) \sim \Pr\) a random pair with a label \(Y \in \YC = \{1, \dots, K \}\) and covariates \(X \in \XC\).
  A probabilistic classifier outputs a vector of predicted class probabilities \(\pmodel(x) = \{\pmodel_1(x), \ldots, \pmodel_K(x)\} \in \Delta_K\), where \(\Delta_K \coloneqq \{q \in [0, 1]^K\colon \sum_{k = 1}^K q_k = 1\}\) is the probability simplex.
  The ground-truth conditional distribution is denoted by \(\ptrue(x)\). 
  When there is no ambiguity, we write
  \(\pmodel \coloneqq \pmodel(x)\) and \(\ptrue \coloneqq \ptrue(x)\). 
  The predicted label is \(\argmax_k \pmodel_k(x)\), and the value \(\max_k \pmodel_k(x)\) is commonly referred to as the \textit{confidence} in the calibration literature~\citep{zhao2021calibrating}.

\subsection{Background on Calibration}

\subsubsection{Notions of Calibration}
  Following~\citet{dwork2021outcome,zhao2021calibrating}, we view calibration from the point of \emph{indistinguishability}.
  A classifier \(\pmodel\) is considered calibrated with respect to some notion of calibration if a model \( \pmodel \) is statistically indistinguishable from the ground truth process \( \ptrue \) with respect to that notion of calibration.

  Ideally, we would like \( \pmodel \) to coincide with \( \ptrue \) for each input \(x\), in which case the model would be perfectly calibrated under any notion of calibration.
  In practice, however, this pointwise equality almost never holds. This has led to many relaxed notions of calibration in the literature.
  One such relaxation is \emph{multiclass (distribution) calibration}.

  \begin{definition}
    \label{def:distribution-calibration}
    A probabilistic classifier \(\pmodel:\mathcal{X}\to\Delta_K\) is said to be \textit{multiclass (distribution) calibrated} if for every \(k\in\YC\) the following equality holds:
    \begin{equation}
        \Pr\bigl(Y = k \mid \pmodel(X) = q\bigr) = q_k.
    \label{eq:distribution-calibration}
    \end{equation}
    \end{definition}

  A multiclass-calibrated classifier can deviate significantly from \(\ptrue\) in a pointwise sense. A simple illustration arises for classifiers with low sharpness~\citep{gruber2022better}.
  Moreover, multiclass calibration is difficult to verify empirically. For problems with many classes and limited data, checking whether~\eqref{eq:distribution-calibration} holds is not feasible, and it has motivated further relaxations with tractable error measures.

  Arguably the most widely used calibration measures are \emph{confidence calibration}~\citep{guo2017calibration} and \emph{class-wise calibration}~\citep{kull2019beyond}.

  \begin{definition}
    A probabilistic classifier \(\pmodel:\mathcal{X}\to\Delta_K\) is said to be \textit{confidence calibrated} if for all \(\beta \in [0,1]\) the following equality holds:
    \begin{equation}
      \Pr\bigl( Y = \argmax_{k\in \YC} \pmodel_k(X) \mid \max_{k\in \YC} \pmodel_k(X) = \beta \bigr) = \beta.
    \label{eq:confidence-calibration}
    \end{equation}
  \end{definition}

  \begin{definition}
    A probabilistic classifier \(\pmodel:\mathcal{X}\to\Delta_K\) is said to be \textit{class-wise calibrated} if for all \(\beta \in [0,1]\) and all \(k\in \YC\)  the following equality holds:
    \begin{equation}
      \Pr\bigl(Y = k \mid \pmodel_k(X) = \beta\bigr) = \beta.
    \label{eq:class-wise-calibration}
    \end{equation}
  \end{definition}

  Both~\eqref{eq:confidence-calibration} and~\eqref{eq:class-wise-calibration} are \emph{necessary} but not \emph{sufficient} for multiclass calibration~\eqref{eq:distribution-calibration}, as it is straightforward to verify.

  Note that all these notions of calibration, by definition, are marginal in the sense that they are not required to hold for a particular \(x\). Rather, they are defined on average for the class of indistinguishable inputs that satisfy the conditions in these definitions; see~\eqref{eq:confidence-calibration} and~\eqref{eq:class-wise-calibration}. E.g., for confidence calibration, we expect that~\eqref{eq:confidence-calibration} holds for all inputs \(x\), for which \( \max_{k\in \YC} \pmodel_k(x) = \beta \) holds true.

\subsubsection{Measures of Calibration Error}
  Unlike multiclass calibration, confidence and class-wise calibration have widely used practical error measures.

  The \textbf{Expected Calibration Error} (ECE; \citealp{guo2017calibration}) evaluates how well a classifier is confidence calibrated:
  \begin{equation*}
    \text{ECE} = \sum_{j=1}^{b} \frac{|B_j|}{N} \bigl|\text{Acc}(B_j)-\text{Conf}(B_j)\bigr|,
  \end{equation*}
  where \(N\) is the size of validation dataset, \(B_j\) is the \(j\)-th bin induced by predicted confidence, \(\text{Acc}(B_j)\) is the accuracy in bin~\(B_j\), and \(\text{Conf}(B_j)\) is the averaged predicted confidence in bin~\(B_j\).

  The \textbf{class-wise Expected Calibration Error} (cw-ECE; \citealp{kull2019beyond}) evaluates class-wise calibration analogously.
  \begin{equation*}
    \text{ECE}_{\text{cw}} = \sum_{k=1}^{K} \sum_{j=1}^{b} \frac{\bigl|B_j^{(k)}\bigr|}{N K} \cdot
    \bigl|\text{Acc}\bigl(B_j^{(k)}\bigr) - \text{Conf}\bigl(B_j^{(k)}\bigr)\bigr|,
  \end{equation*}
  where each bin \(B_j^{(k)}\) is formed separately for each class.

  In literature, several post-hoc calibration algorithms have been proposed to reduce these errors. These include (non-adaptive) temperature scaling~\citep{guo2017calibration} for ECE and Dirichlet calibration~\citep{kull2019beyond} for cw-ECE.
  However, despite showing practical improvements in some cases, these algorithms do not, in general, reduce the corresponding error measure.

  In what follows, we introduce a new set-based notion of calibration together with an algorithm that is guaranteed to reduce the corresponding calibration error and provides distribution-free marginal coverage guarantees.

\subsection{Cumulative Mass Calibration}
\label{sec:mass_calibration}

\textbf{Practical motivation.}
  We start this section with the motivation for the set-based calibration. Each notion of calibration helps different downstream decision-making rules~\citep{zhao2021calibrating}.
  For example, confidence calibration aligns predicted confidence with empirical accuracy, leading to reliable selective prediction, abstention, or error detection.
  Class-wise calibration reduces gaps between predicted class probabilities and actual class frequencies, thereby lowering class-specific over- or underconfidence.
    
  In many applications, however, systems operate with \emph{prediction sets} rather than single labels~\citep{Sadinle02012019, NEURIPS2020_Romano, DBLP:journals/corr/abs-2009-14193}. 
  For example, such systems might collect top-ranked labels until a target probability mass is reached. Based on the number of classes needed to get the target mass, the system decides whether to proceed. Then, it can proceed only if the resulting set of classes is sufficiently small.
  For decisions based on these sets to be reliable, the cumulative mass of predicted probabilities must match the empirical coverage of the true labels.
  This motivates us to introduce new calibration notions, defined for cumulative masses.

\textbf{Definitions.}
  Let \(q\) be any categorical distribution over \(K\) classes and let \(\pi\) be a permutation of indices such that \(q_{\pi(1)}, \dots, q_{\pi(K)}\) is nonincreasing.
  The \emph{highest-probability region} of mass \(1 - \alpha\) is:
  \begin{equation*}
    \HPRa[q] \coloneqq \{\pi(1), \ldots, \pi(m)\},
  \end{equation*}
  where \(m \coloneqq \min\bigl\{r\colon \sum_{k=1}^{r} q_{\pi(k)} \ge 1 - \alpha\bigr\}\).

  Now, let us introduce two new notions of calibration.
  \begin{definition}
    A probabilistic classifier \(\pmodel:\mathcal{X}\to\Delta_K\) is said to be \textit{\(\alpha\)-cumulative mass calibrated} if for a given \(\alpha \in [0,1]\)  the following inequality holds:
    \begin{equation*}
      \Pr\bigl(Y \in \HPRa[\pmodel(X)]\bigr) \ge 1 - \alpha.
    \end{equation*}
    \label{def:alpha_mass_calibration}
  \end{definition}
  Importantly, this condition holds for the ground-truth classifier \(\ \ptrue \) (see Appendix~\ref{sec:appendix_definition_check}).
  Equality need not hold, even for \(\ptrue\), because the last included class may cause the cumulative mass to overshoot \(1 - \alpha\).

  \begin{definition}
    A probabilistic classifier is said to be \textit{cumulative mass calibrated} if it is \(\alpha\)-cumulative mass calibrated for every \(\alpha \in [0,1]\).
    \label{def:mass_calibration}
  \end{definition}
  Again, this condition holds for the ground-truth distribution \(\ptrue\) (see Appendix~\ref{sec:appendix_definition_check}). 

  Unlike notions that calibrate only a single class probability or a confidence score, which operate with scalars, cumulative mass calibration requires calibration of a \emph{set of predicted probabilities}.

\subsection{Cumulative Mass Calibration Errors~(CMCE)}
\label{sec:cmce}
  In this section, we introduce measures of calibration error that, analogously to ECE and cw-ECE, assess how well a classifier is calibrated with respect to the calibration notions defined above.

\textbf{\(\boldsymbol{\alpha}\)-CMCE.}
  We first define the \emph{empirical coverage} of a model \(\pmodel\) on a validation set \(\DC_{val}\) of size \(N\) as the fraction of examples whose true label falls in the highest-probability region \(\HPRa(\pmodel)\):
  \begin{equation*}
    \text{Coverage}(\pmodel, \DC_{val}) = \frac{1}{N} \sum_{i=1}^N \indicator\bigl(y_i \in \HPRa[\pmodel(x_i)] \bigr).
  \end{equation*}
  The \(\alpha\)-CMCE is the deviation of coverage from the target coverage level \(1-\alpha\):
  \begin{equation}
    \alpha\text{-CMCE}(\pmodel, \DC_{val}) = |\text{Coverage}(\pmodel, \DC_{val}) - (1 - \alpha)|.
  \label{eq:cmce-def}
  \end{equation}

\textbf{CMCE.}
  Let us introduce CMCE, another measure of calibration error that does not require specifying \(\alpha\) for computation. 
  This measure is computed using binning, similarly to ECE and cw-ECE. 
  For a validation set of size \(N\), we form, for each example \(x\), the \(K\) nested highest-probability sets induced by sorting the predicted probabilities.
  All such sets are collected into \(\SC\) (giving \(|\SC| = K N\)), and the mass axis is partitioned into \(b\) bins with edges \(0 = t_0 < t_1 < \cdots < t_b = 1\).
  Each set \(S \in \SC\) is assigned to bin \(j\) according to its cumulative mass. The number of bins \(b\) is a hyperparameter, chosen as in ECE and cw-ECE.

  Let \(\SC_j\) denote the collection of sets in bin \(j\).
  Within each bin, we compute two quantities. 
  The \emph{empirical coverage} \(\mathrm{cov}(\SC_j)\) is the fraction of sets in \(\SC_j\) that contain the true label \(y\). 
  The \emph{average cumulative mass} \(\mathrm{mass}(\SC_j)\) is the mean mass of sets in \(\SC_j\).
  The cumulative mass calibration error is
  \begin{equation}
    \text{CMCE} = \frac{1}{|\SC|} \sum_{j=1}^b |\SC_j| \cdot |\text{cov}(\SC_j) - \text{mass}(\SC_j)|,
  \label{eq:cmce_definition}
  \end{equation}
  where \(|\SC_j|\) is the number of sets in bin \(j\).
  Since both coverage and mass lie in \([0, 1]\), we have \(0 \le \text{CMCE} \le 1\), with smaller values indicating better calibration.

  In the next section, we introduce a calibration algorithm that minimizes \(\alpha\)-CMCE and provides \(\alpha-\)cumulative mass calibrated classifier (see Definition~\ref{def:alpha_mass_calibration}).
  While we cannot provide analogous guarantees for CMCE, our experiments show that the algorithm also yields empirical improvements in CMCE.


\section{Mass Calibration Algorithms}
\label{sec:method}
  In this section, we introduce several algorithms that optimized the proposed variants of cumulative mass calibration.

\subsection{Direct CMCE minimization}
  We start with the most straightforward approach of direct CMCE optimization with \emph{non-adaptive} temperature scaling.
  In its classical form~\citep{guo2017calibration}, temperature scaling selects a single \emph{global} temperature by maximizing predictive likelihood on a calibration set.
  For direct CMCE minimization, we instead select the temperature by minimizing CMCE (empirically) on a hold-out set. 
  Since CMCE is not differentiable, we fix a predefined grid of temperature values and select the optimal one (see details in Appendix~\ref{sec:appendix_naive_cmce}).

  The benefit of this algorithm, which we refer to as \emph{Naive CMCE}, is that it optimizes CMCE directly, simultaneously for all levels \( \alpha \).
  However, it relies on a search over a non-differentiable objective and provides no guarantee that the resulting classifier satisfies the corresponding notion of calibration (see Definition~\ref{def:mass_calibration}). 
  To obtain the guarantees, we construct an algorithm based on Conformal Prediction (CP) to produce an \(\alpha\)-cumulative-mass-calibrated classifier.
  CP provides a marginal coverage guarantee for the produced sets. Using the conformal set \(\CC_\alpha(X)\), we design a procedure that enforces the mass constraint at the chosen level \(\alpha\). 
  We next review CP and then present the algorithm.

\subsection{Background on Conformal Prediction}
\label{sec:background_conformal}
  Conformal prediction~\citep{shafer2008tutorial,angelopoulos2021gentle} is a framework that, under the exchangeability assumption, constructs a \emph{set of labels} \(\CC_\alpha(X)\) with a \emph{marginal} coverage guarantee:
  \begin{equation}
    \Pr \bigl(Y \in \CC_\alpha(X)\bigr) \ge 1-\alpha,
  \label{eq:marginal-coverage}
  \end{equation}
  for any specified error level \(\alpha \in (0, 1)\).

  To construct a CP label set (in a split conformal setup), one needs a pretrained model and a nonconformity score \(s \colon \XC \times \YC \to \RR\) that measures the error between a prediction and the correct target.
  The procedure requires a threshold \(q_{\alpha}\), selected from the scores \(\{ s(x_i,y_i) \}_{i=1}^n\) on a \emph{calibration dataset}.
  The threshold is defined as the \(\ceil{(1 - \alpha)(n + 1)} / n \) quantile of the empirical score distribution.
  Once the quantile is selected, the prediction set for an input \( x \) is formed as \(\CC_\alpha(x) \coloneqq \{y\colon s(x, y) \le q_{\alpha}\}\).

  However, the coverage guarantee in~\eqref{eq:marginal-coverage} holds only \emph{marginally} and says nothing about a set constructed for a specific input \(x\).
  Ideally, one would like to have \emph{a conditional} coverage guarantee \( \Pr(Y \in \CC_\alpha(X) \mid X = x) \ge 1 - \alpha \) for any \( x \) and any \( \alpha \). 
  However, this is impossible without additional distributional assumptions~\citep{vovk2012conditional,foygel2021limits}.
  Nevertheless, the following thoughtful procedure demonstrates the link between conditional conformal sets and perfect (point-wise) calibration.

  Suppose that for every \( x \) and any confidence level \( \alpha \), we have valid conditional conformal sets \( \{\CC_\alpha(x)\}_{\alpha \in (0,1)} \). Now let us fix an input \(x\).
  Consider two boundary confidence levels \(\alpha_1 \leq \alpha \leq \alpha_2\) such that the corresponding sets differ by exactly one label \(y\). Specifically, this means \(\CC_{\alpha_1}(x) = \CC_{\alpha_2}(x)\cup\{y\}\). Assume additionally that both \( \alpha_1 \) and \( \alpha_2 \) are minimal for the label sets they correspond to, and that the coverage is tight at these two levels.
  We can then recover the ground-truth probability of \(y\) as
  \begin{equation*}
    \begin{split}
      p(Y = y \mid X = x) &= \\
      \Pr\bigl(Y \in \CC_{\alpha_1}(X) \mid X = x\bigr)
      &- \Pr\bigl(Y \in \CC_{\alpha_2}(X) \mid X = x\bigr) \\
      &= \alpha_2 - \alpha_1.
    \end{split}
  \end{equation*}
  If we repeat this procedure for every label (by varying corresponding boundary levels \(\alpha\)), we can recover the full conditional distribution \(p(y \mid x)\) and reconstruct the ground-truth process for every \(x\).
  %

  This preceding example demonstrates that valid (and exact at the boundaries) conditional sets can determine the full conditional distribution in~\eqref{eq:distribution-calibration}. Hence, conformal prediction can be used to improve calibration.

  However, these valid conditional sets are unavailable, and in practice, we can only construct marginally valid sets \(\CC_\alpha(X)\) using conformal prediction.
  In the next section, we present a calibration algorithm based on this idea.

\subsection{Cumulative Mass Calibration via CP}
\label{sec:mass_rescaling}
  We propose a calibration algorithm, based on a marginal conformal prediction, with \textit{adaptive} temperature scaling \emph{for each input \(x\)}.
  The algorithm is \emph{fundamentally different} from classical temperature scaling in~\citep{guo2017calibration}, which fits a single global temperature shared across all inputs.
  Our method tunes an \emph{input-dependent} temperature \(\tau(x)\) which can be effectively found via one-dimensional binary search applied to a monotone function.

  Below, we assume that both the pretrained model \( \pmodel \) and the procedure for constructing conformal sets are given.
  At inference time, using the uncalibrated model and its corresponding fitted CP procedure for specified level \( \alpha \), we produce a conformal set for input \(x\).
  Then, we scale the model's raw logits by the temperature \(\tau(x) > 0\), which is tuned to satisfy a cumulative mass constraint for each input \(x\).
  Once this temperature \(\tau(x)\) is found, the classifier is calibrated in a sense that the cumulative mass over \(\CC_\alpha(X)\) matches (or, when not possible to match, exceeds) \(1-\alpha\). 
  The detailed procedure is given in Algorithm~\ref{alg:conformal_temperature_scaling}.

  In what follows, we omit explicit dependency on \(x\) for temperature: \(\tau\coloneqq\tau(x)\).
  The optimization problem for each input \(x\) is as follows:
  \begin{equation*}
    \begin{aligned}
      \min_{\tau > 0}\quad & \sum_{y \in \CC_\alpha(x)} \tilde{p}(x, \tau)_y, \\
      \text{s.t.} \quad & \sum_{y \in \CC_\alpha(x)} \tilde{p}(x, \tau)_y \ge 1 - \alpha,
    \end{aligned}
  \label{eq:closest_from_above}
  \end{equation*}
  where \(\tilde{p}(x, \tau) = \text{Softmax}\bigl[\log \pmodel(x) / \tau\bigr]\).

  The problem is one-dimensional for a scalar \( \tau \). Since the cumulative mass over \(\CC_\alpha(x)\) changes monotonically with \( \tau \), it can be solved via one-dimensional binary search.

\newcommand{\AlgoSection}[1]{%
\noindent\rule{0.90\linewidth}{0.01pt}\par
  \noindent\textbf{\normalsize #1}\par\vspace{0.0em}%
}

\begin{algorithm}[t]
\caption{\textsc{Conformal Temperature Scaling}}
\label{alg:conformal_temperature_scaling}
\DontPrintSemicolon
\footnotesize
\LinesNotNumbered 

\KwIn{%
\begin{tabular}[t]{@{}l@{}}
Calibration set $\mathcal{D}_{\mathrm{cal}}=\{(x_i,y_i)\}_{i=1}^n$; \\
target error level $\alpha\in(0,1)$; \\
probabilistic predictor $\widehat{p}\colon \mathcal{X}\to\Delta^{K-1}$; \\
conformal prediction hyperparameters $s$, $f_\theta$; \\
temperature bounds $0<\tau_{\min}<\tau_{\max}$.
\end{tabular}%
}
\KwOut{Calibrated probability vector $q(x)\in\Delta^{K-1}$.}

\AlgoSection{Calibrator fitting stage:}

 \textbf{Initialize conformal predictor} $\mathrm{CP}$ with hyper-parameters: $\alpha$, $s$, $f_\theta$. And \textbf{fit} it on $\mathcal{D}_{\mathrm{cal}}$, processed with $\hat{p}$.

\AlgoSection{Calibration stage:}

\nl \textbf{Conformal predictor call.} For a test input $x$, compute
\[
C_\alpha(x) \leftarrow \mathrm{CP.predict}\bigl(\widehat{p}(x)\bigr).
\]

\nl \If{$C_\alpha(x)=\varnothing$ \textbf{or} $|C_\alpha(x)|=K$}{
    \Return{$\widehat{p}(x)$}.\;
}

\nl \textbf{Temperature-scaled probabilities.} For any $\tau>0$, define
\begin{equation*}
\begin{aligned}
\bar{p}_x(\tau)
&= \sum_{y\in C_\alpha(x)}
\mathrm{softmax}\!\left(\frac{\log \hat{p}(x)}{\tau}\right)_y
&= \sum_{y \in \mathcal{C}_\alpha(x)}
\tilde{p}(x, \tau)_y .
\end{aligned}
\end{equation*}

\nl \textbf{Temperature selection.} Find
\[
\tau^\star
\in
\arg\min_{\tau\in[\tau_{\min},\tau_{\max}]}
\bar{p}_x(\tau)
\quad, \quad
\text{s.t.}
\quad
\bar{p}_x(\tau)\ge 1-\alpha ,
\]
using one-dimensional  binary search over $\tau$.\;

\BlankLine
\nl \Return{$q=\tilde{p}(x, \tau)$}.\;
\end{algorithm}

\subsection{Construction of Predictive Sets}
\label{sec:construction_of_predictive_sets}
  The proposed procedure linking valid conditional coverage to point-wise calibration (see Section~\ref{sec:background_conformal}) cannot be realized without additional assumptions~\citep{vovk2012conditional}.
  Therefore, in practice, for our \emph{adaptive} temperature scaling algorithm, we use predictive sets from (split) CP, which provide only marginal (not conditional) coverage guarantees.
  Such marginally valid sets can be conservative, especially when per-class error rates are heterogeneous. As a result, a single global threshold may be too strict in some regions of \(\XC\).

  Recent works propose adaptive \emph{threshold} selection (do not confuse with adaptive temperature scaling) procedures that transform nonconformity scores in an \(x\)-dependent, invertible way while preserving marginal coverage~\citep{colombo2024normalizing,plassierrectifying}.
  These techniques also keep marginal coverage guarantees while often empirically improving conditional coverage.
  More precisely, let us define a transformation of the nonconformity score as:
  \begin{equation}
    \tilde{s}(x, y) = f_{\theta}(s(x, y), x),
  \label{eq:instance_dependent_score}
  \end{equation}
  where \(f_{\theta}\colon \RR \times \XC \to \RR\) is an invertible function parameterized by \(\theta\).
  We select a global threshold \(q_\alpha\) for the transformed scores as in Section~\ref{sec:background_conformal}, and then define an \(x\)-adaptive threshold \(\tilde{q}_{\alpha}(x) = f^{-1}_{\theta}(q_\alpha, x)\).

  Accordingly, we consider two strategies for threshold selection in conformal prediction:
  \begin{enumerate}
    \item \emph{Constant}: the classical global threshold \(q_\alpha\) as in Section~\ref{sec:background_conformal}, with no additional training.

    \item \emph{Adaptive}: learn an invertible \(f_\theta\), select a global threshold based on the transformed scores, and adapt per \(x\) via \(\tilde{q}_{\alpha}(x) = f^{-1}_{\theta}(q_\alpha, x)\). In our experiments, we use the method of~\citet{colombo2024normalizing}.
  \end{enumerate}

\subsection{Properties of Calibrated Classifiers}
\label{sec:properties}
  Our calibrated classifiers preserve the \emph{marginal} coverage guarantee from conformal prediction at the chosen level \(\alpha\), since by construction \(\Pr\bigl(Y \in \CC_\alpha(X)\bigr) \ge 1 - \alpha\). 
  Given the conformal set \(\CC_\alpha(x)\), our \emph{adaptive} temperature scaling keeps the class ordering unchanged (due to monotonicity of temperature scaling) and tunes a single scalar temperature for that input.
  
  Hence, after the calibration procedure, the probability mass within a predicted conformal set meets \(1-\alpha\) when feasible, and otherwise exceeds it. In particular, $\HPRa[(\tilde{p}(x, \tau)] = \CC_\alpha(x)$ and  $\Pr\bigl(Y \in \HPRa[(\tilde{p}(x, \tau)]\bigr) \ge 1 - \alpha$.  Moreover, the accuracy also remains the same.
  The mentioned marginal coverage guarantees hold only for the selected \(\alpha\).
  Therefore, the adaptive temperature scaling algorithm ensures \(\alpha\)-cumulative mass-calibration of a classifier (see Definition~\ref{def:alpha_mass_calibration}).
  
  In contrast to direct CMCE fitting, adaptive temperature scaling does not imply calibration for other \(\alpha\) values (although in practice there could be improvements; see Section~\ref{sec:experiments}).
  As the CMCE error measure summarizes calibration across all cumulative masses, we treat it as a \emph{diagnostic}.
  While we do not guarantee CMCE improvement, we often observe empirical improvements when we enforce the conformal mass constraint.


  \begin{figure*}[!htb]
    \centering
    \begin{subfigure}[t]{0.85\linewidth}
      \centering
      \includegraphics[width=\linewidth]{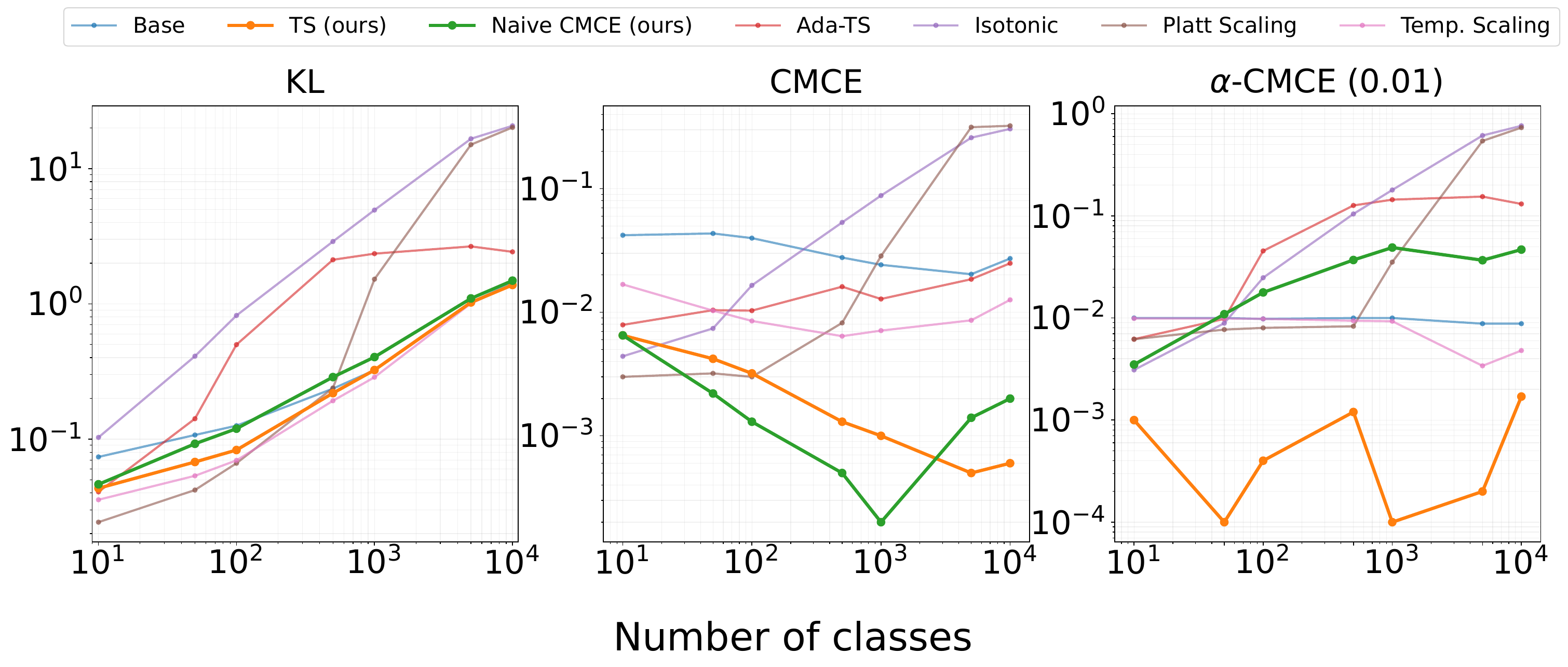}
    \end{subfigure}%
    \caption{Performance metrics on synthetic data, as the number of classes \(K\) increases. TS tuned for \( \alpha=0.01 \), MSP nonconformity score, and Constant transformation. KL computed with ground-truth \( \ptrue(y \mid x) \). Our method (TS) consistently achieves better metrics as \(K\) increases.}
  \label{fig:synthetic_experiment_results}
  \end{figure*}

\section{Experiments}
\label{sec:experiments}
  In this section, we evaluate proposed calibration methods in classification problems. 
  We first consider a synthetic experiment in which the ground-truth probabilities \( \ptrue \) are known. We then consider real-world image classification datasets with a varying number of classes. 
  As we will see, a large number of classes pose a particularly challenging calibration problem for many calibration methods.
    
  Empirically, we observe that once a classifier is \( \alpha \)-cumulative-mass calibrated, it often also improves CMCE, ECE~\citep{guo2017calibration}, cw-ECE~\citep{kull2019beyond}, and other popular calibration metrics.

\textbf{Baselines.}
  We compare our methods with popular calibration methods: Isotonic Regression~\citep{zadrozny2002transforming}, Platt Scaling~\citep{platt1999probabilistic}, Global Temperature Scaling~\citep{guo2017calibration}, Adaptive Temperature Scaling (Ada-TS; \citealp{joy2023sample}), Dirichlet Scaling~\citep{kull2019beyond},  and Venn-Abers calibration~\citep{vovk2014vennaberspredictors}.
  We describe the baselines in Appendix~\ref{sec:appendix_baseline_description}.

\textbf{Our method.}
  We propose a calibrator, \texttt{TS (ours)}, that refers to \emph{conformal temperature scaling}, shown in the flow chart in Figure~\ref{fig:flow_chart} and presented in Algorithm~\ref{alg:conformal_temperature_scaling}.
  %

  %
  Since the algorithm is based on conformal prediction sets, one needs to choose a nonconformity score. We consider two options: \texttt{APS} (Adaptive Predicted Sets; \citealp{NEURIPS2020_Romano}), defined as \( s_{\text{APS}}(x, y) = \sum_{j=1}^{k} \pmodel_{\pi(j)}(x) \) where \( y = \pi(k) \), and \texttt{MSP} (Maximum Softmax Probability), defined as \( s_{\text{MSP}}(x, y) = 1 - \pmodel_y(x) \).

  Finally, we consider two options for transforming the threshold. One can use a standard (split) CP procedure with a single \emph{global threshold} (see Section~\ref{sec:background_conformal}) or an optional transformation that allows instance-dependent thresholds, as given by equation~\eqref{eq:instance_dependent_score}. 

  In our notation, we denote these choices by \texttt{Const} (Constant), the case when no threshold transformation is applied, and by \texttt{Ada} (Adaptive), the case when a normalizing-flow transformation is used~\citep{colombo2024normalizing}.

\subsection{Synthetic Experiment}
  We evaluate calibration on a synthetic dataset where the ground-truth probabilities \( \ptrue( y \mid x) \) are available, which allows us to compute values of Kullback-Leibler divergence (KL) exactly, using \( \ptrue \). Note that KL is a proper calibration error (it is 0 iff \( \ptrue = \pmodel \) for all \(x\))~\citep{gruber2022better}, and the smaller it is, the better calibration is achieved.
  A complete description of the data-generating process is presented in Appendix~\ref{sec:appendix_synthetic_experiment}. Figure~\ref{fig:synthetic_experiment_results} summarizes the main results.

  The proposed algorithms (Naive CMCE and TS) demonstrate strong performance across different metrics and numbers of classes \( K \).
  Moreover, our \emph{adaptive} TS (instantiated with constant score transformation, MSP scores, and \( \alpha=0.01 \)) consistently achieves low KL divergence and substantially reduces CMCE, despite not being explicitly optimized for it.
  Our adaptive TS is the only calibration procedure that does not degrade \(\alpha\)-CMCE with the increase of \( K \). 

  The synthetic experiment clearly shows that our \emph{adaptive} TS calibration method shows  reasonable results for a moderate number of classes, while significantly improving performance when the number of classes is large. We observe the same behaviour in real-world image datasets.

  \begin{figure}[!t]
    \centering
    \includegraphics[width=0.8\linewidth]{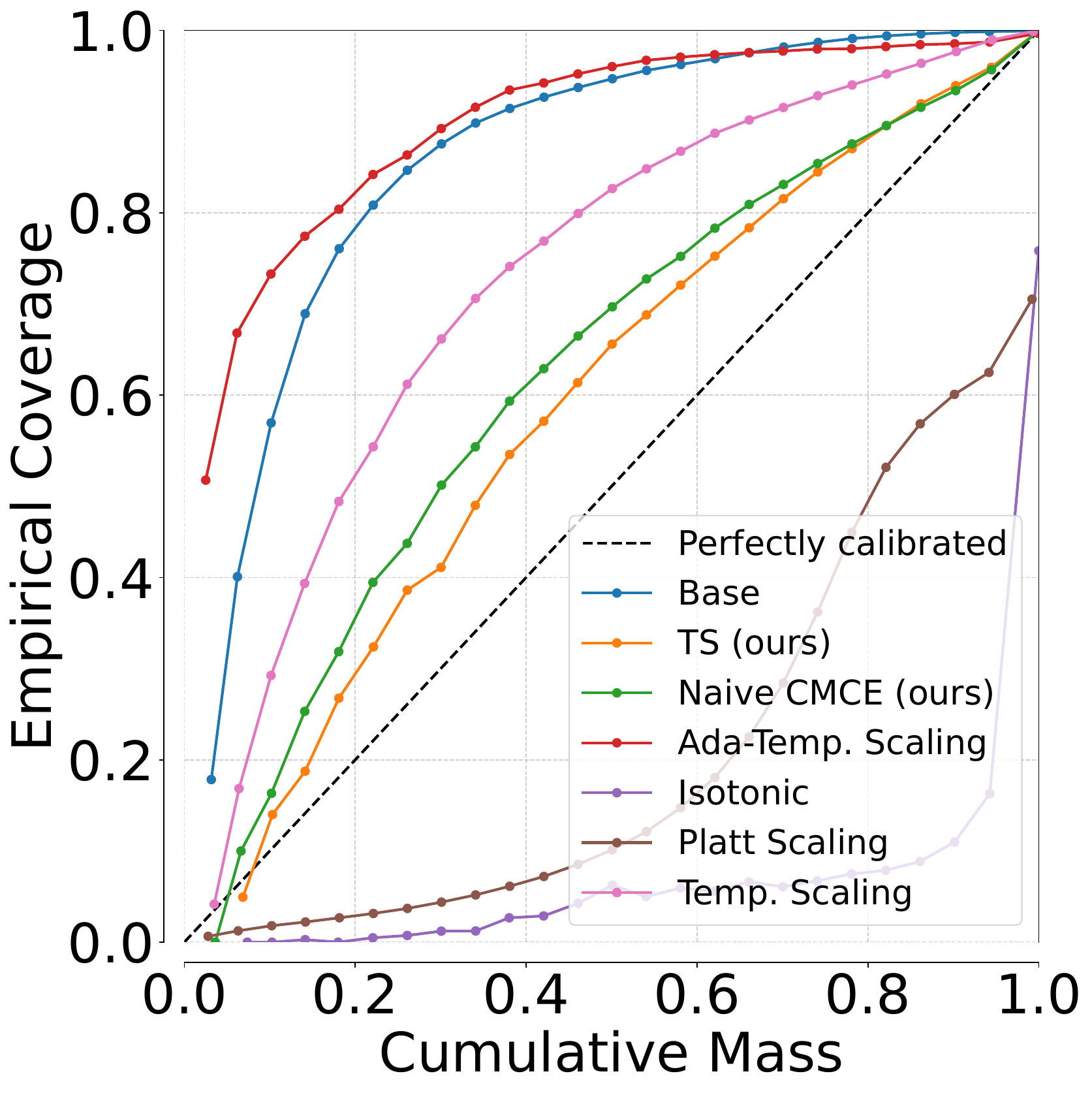}
    \caption{Cumulative mass calibration curves for different calibrators on iNaturalist21.}
  \label{fig:cum_mass_two_datasets}
  \end{figure}

\subsection{Image Classification Datasets}
  In this section, we consider popular visual datasets and study the calibration of image classifiers.
  We consider datasets with different numbers of classes, such as CIFAR-10, CIFAR-100~\citep{krizhevsky2009learning}, ImageNet~\citep{imagenet} with 1,000 classes, and iNaturalist~\citep{vanhorn2021benchmarkingrepresentationlearningnatural}, a dataset with a large (10,000) number of classes. 
  More detailed description of the datasets is given in Appendix~\ref{sec:appendix_datasets_description}.
  For all datasets, we use pretrained, publicly available models trained in a standard manner (without train-time methods to improve calibration).
  We provide a detailed description of the models we use in Appendix~\ref{sec:appendix_models_description}. 
  Due to computational constraints, we do not evaluate some baselines on particular datasets. We specify which baselines are omitted for each dataset and provide details in~\ref{sec:appendix_computational_constraits}.

  \begin{table*}[htb]
\centering
\captionsetup{justification=centering}
\caption{\textbf{Upper table}: CIFAR-100 with ResNet-56; \textbf{Lower table}: iNaturalist21. 
Our calibrators use the MSP nonconformity score, a constant transformation, and either \(\alpha = 0.05\) (upper) or \(\alpha = 0.1\) (lower).
Metrics\textsuperscript{*} are scaled by \(100\). \textcolor{red}{Orange}/\textcolor{ForestGreen}{Green} indicate worse/better performance than the uncalibrated model. \textbf{Bold} and \underline{underline} mark the best and second-best values.
}
\label{tab:results_cifar100_inaturalist_best}
\resizebox{0.89\textwidth}{!}{%
\begin{tabular}{@{}l @{\hspace{8pt}}c c c c c c c}
\toprule
\multicolumn{7}{c}{\textbf{CIFAR-100}} \\
\midrule
Calibrator & ECE\textsuperscript{*} & MCE\textsuperscript{*} & NLL & Brier Score\textsuperscript{*} & CMCE\textsuperscript{*} & $\alpha$-CMCE (0.05)\textsuperscript{*} \\
\midrule
Base
& $14.17 \spm 0.10$ & $32.36 \spm 1.89$ & $1.29 \spm 0.01$ & $41.78 \spm 0.16$ & $1.29 \spm 0.01$ & $8.09 \spm 0.12$ \\
\midrule
Ada-TS
& \textcolor{red}{$15.04 \spm 0.24$} & \textcolor{ForestGreen}{$29.45 \spm 2.93$} & \textcolor{red}{$2.37 \spm 0.05$} & \textcolor{red}{$44.80 \spm 0.26$} & \textcolor{red}{$1.95 \spm 0.14$} & \textcolor{red}{$8.75 \spm 0.25$} \\
Dirichlet
& \textcolor{red}{$40.77 \spm 1.01$} & \textcolor{red}{$73.60 \spm 3.71$} & \textcolor{red}{$10.44 \spm 0.55$} & \textcolor{red}{$82.06 \spm 1.94$} & \textcolor{red}{$3.41 \spm 0.17$} & \textcolor{red}{$35.76 \spm 1.05$} \\
Isotonic
& \textcolor{ForestGreen}{$5.93 \spm 0.29$} & \textcolor{red}{$32.52 \spm 33.85$} & \textcolor{red}{$1.73 \spm 0.07$} & \textcolor{ForestGreen}{$39.28 \spm 0.19$} & \textcolor{red}{$1.73 \spm 0.16$} & \textcolor{ForestGreen}{$2.27 \spm 0.33$} \\
Platt Scaling
& \textcolor{red}{$14.56 \spm 0.23$} & \textcolor{ForestGreen}{$21.73 \spm 0.80$} & \textcolor{ForestGreen}{$1.14 \spm 0.01$} & \textcolor{ForestGreen}{$41.72 \spm 0.11$} & \textcolor{red}{$1.46 \spm 0.06$} & \textcolor{ForestGreen}{$3.42 \spm 0.09$} \\
Temp. scaling
& \textcolor{ForestGreen}{$\mathbf{2.85 \spm 0.15}$} & \textcolor{ForestGreen}{$\underline{11.41 \spm 2.81}$} & \textcolor{ForestGreen}{$\mathbf{1.04 \spm 0.01}$} & \textcolor{ForestGreen}{$\underline{38.24 \spm 0.13}$} & \textcolor{ForestGreen}{$0.45 \spm 0.03$} & \textcolor{ForestGreen}{$1.42 \spm 0.08$} \\
V.-Abers (OvA)
& \textcolor{ForestGreen}{$8.14 \spm 0.44$} & \textcolor{ForestGreen}{$16.67 \spm 1.48$} & \textcolor{ForestGreen}{$1.12 \spm 0.01$} & \textcolor{ForestGreen}{$39.36 \spm 0.22$} & \textcolor{red}{$3.84 \spm 0.15$} & \textcolor{ForestGreen}{$4.44 \spm 0.08$} \\
\midrule
Naive CMCE (ours)
& \textcolor{ForestGreen}{$\underline{3.92 \spm 0.42}$} & \textcolor{ForestGreen}{$\mathbf{11.17 \spm 0.90}$} & \textcolor{ForestGreen}{$1.04 \spm 0.01$} & \textcolor{ForestGreen}{$38.33 \spm 0.12$} & \textcolor{ForestGreen}{$\mathbf{0.12 \spm 0.08}$} & \textcolor{ForestGreen}{$\underline{0.57 \spm 0.25}$} \\
TS (ours)
& \textcolor{ForestGreen}{$4.26 \spm 0.33$} & \textcolor{ForestGreen}{$11.99 \spm 0.94$} & \textcolor{ForestGreen}{$\underline{1.04 \spm 0.01}$} & \textcolor{ForestGreen}{$\mathbf{38.15 \spm 0.16}$} & \textcolor{ForestGreen}{$\underline{0.35 \spm 0.02}$} & \textcolor{ForestGreen}{$\mathbf{0.33 \spm 0.21}$} \\
\bottomrule

\toprule
\multicolumn{7}{c}{\textbf{iNaturalist21}} \\
\midrule
Calibrator & ECE\textsuperscript{*} & MCE\textsuperscript{*} & NLL & Brier Score\textsuperscript{*} & CMCE\textsuperscript{*} & $\alpha$-CMCE (0.1)\textsuperscript{*} \\
\midrule
Base
& $49.73 \spm 0.29$ & $64.52 \spm 0.82$ & $2.04 \spm 0.01$ & $61.49 \spm 0.26$ & $5.55 \spm 0.02$ & $9.61 \spm 0.04$ \\
\midrule
Ada-TS
& \textcolor{ForestGreen}{$\underline{11.48 \spm 0.53}$} & \textcolor{ForestGreen}{$26.97 \spm 1.60$} & \textcolor{red}{$2.24 \spm 0.10$} & \textcolor{ForestGreen}{$35.98 \spm 0.47$} & \textcolor{ForestGreen}{$0.85 \spm 0.12$} & \textcolor{ForestGreen}{$\underline{3.03 \spm 0.72}$} \\
Isotonic
& \textcolor{ForestGreen}{$26.43 \spm 1.27$} & \textcolor{ForestGreen}{$38.47 \spm 1.40$} & \textcolor{red}{$13.71 \spm 0.12$} & \textcolor{red}{$80.19 \spm 1.43$} & \textcolor{red}{$17.54 \spm 0.60$} & \textcolor{red}{$40.01 \spm 0.48$} \\
Platt Scaling
& \textcolor{ForestGreen}{$13.97 \spm 0.45$} & \textcolor{ForestGreen}{$29.66 \spm 0.48$} & \textcolor{red}{$12.51 \spm 0.09$} & \textcolor{red}{$68.02 \spm 0.31$} & \textcolor{red}{$24.59 \spm 0.27$} & \textcolor{red}{$32.98 \spm 0.32$} \\
Temp. scaling
& \textcolor{ForestGreen}{$13.16 \spm 8.35$} & \textcolor{ForestGreen}{$\underline{25.60 \spm 11.04}$} & \textcolor{ForestGreen}{$\underline{1.29 \spm 0.10}$} & \textcolor{ForestGreen}{$35.70 \spm 3.76$} & \textcolor{ForestGreen}{$0.28 \spm 0.35$} & \textcolor{ForestGreen}{$4.64 \spm 1.86$} \\
\midrule
Naive CMCE  (ours)
& \textcolor{ForestGreen}{$12.70 \spm 3.73$} & \textcolor{ForestGreen}{$25.73 \spm 5.24$} & \textcolor{ForestGreen}{$\mathbf{1.27 \spm 0.03}$} & \textcolor{ForestGreen}{$\underline{35.09 \spm 1.13}$} & \textcolor{ForestGreen}{$\underline{0.13 \spm 0.01}$} & \textcolor{ForestGreen}{$4.75 \spm 0.95$} \\
TS (ours)
& \textcolor{ForestGreen}{$\mathbf{7.45 \spm 0.13}$} & \textcolor{ForestGreen}{$\mathbf{20.52 \spm 8.27}$} & \textcolor{ForestGreen}{$1.38 \spm 0.04$} & \textcolor{ForestGreen}{$\mathbf{32.73 \spm 0.45}$} & \textcolor{ForestGreen}{$\mathbf{0.08 \spm 0.01}$} & \textcolor{ForestGreen}{$\mathbf{0.18 \spm 0.53}$} \\
\bottomrule
\end{tabular}
    }
\end{table*}

\textbf{Experimental setup.}
  Using available pretrained models for each dataset, we compute predictions (in the form of logits) for the corresponding validation data.
  Then, we split these predictions into two subsets: \textit{calibration} (20\%, to fit calibrators) and \textit{test} (80\%). 
  We use 10 different splits to compute statistics for the metrics we consider.
  Once the calibrators are fitted to the corresponding calibration splits, we evaluate them over the test split and report the metrics.

\textbf{Experimental results.}
  Due to the space constraints, we move some of the results to Appendix~\ref{sec:appendix_additional_experimental_results}. 
  Here, we keep the most interesting ones. 
  Since on some datasets (CIFAR-10, CIFAR-100, and ImageNet) all calibration methods keep the same accuracy as the base model, we do not report it in the tables. 
  However, for iNaturalist, some baselines degrade accuracy, and we report this in a dedicated column in the table below.

\textbf{CIFAR-10.}
  It is well known~\citep{kapoor2022uncertainty} that standard image-classification datasets (e.g., CIFAR-10/100) contain almost no aleatoric uncertainty.
  Therefore, a model that predicts probability vectors close to one-hot vectors tends to be well-calibrated.
  Tables~\ref{tab:results_cifar10_resnet20_best} and~\ref{tab:results_cifar10_resnet56_best} in Appendix~\ref{sec:appendix_additional_experimental_results} reflect this: even uncalibrated models already perform strongly across most metrics.
  We observe that all calibration baselines and our proposed algorithms outperform the uncalibrated baselines.
  Among calibrators, Isotonic Regression and Temperature Scaling typically perform best.
  However, for \( \alpha \)-CMCE, our algorithm with a constant transformation and \( \alpha = 0.01\) gives, as expected, the best result.
  Nevertheless, the setting with few classes and minimal aleatoric uncertainty is not where our methods yield the most significant gain.

\textbf{CIFAR-100.}
  Table~\ref{tab:results_cifar100_inaturalist_best} (upper) reports results for CIFAR-100. 
  In this and subsequent tables, entries that outperform the uncalibrated baseline \textcolor{ForestGreen}{are colored in green}, while those that underperform \textcolor{red}{are colored in orange}.
  From Table~\ref{tab:results_cifar100_inaturalist_best} (upper), we see that our approaches consistently improve all metrics relative to the uncalibrated baseline, achieving the best and second-best results for the CMCE and \(\alpha\)-CMCE. 
  The test accuracy on CIFAR-100 is the same across all calibration methods and is \(0.7264 \pm 0.0023\) for ResNet56 (the model we used). 

\textbf{iNaturalist21.}
  Additionally, we consider the dataset with the largest number of classes (10,000).
  Table~\ref{tab:results_cifar100_inaturalist_best} (lower) reports different test metrics.
  From this table, we see that methods that aggregate binary calibrators into a multiclass calibrator become less effective and often worsen metrics.
  For instance, with Isotonic Regression, we observe a significant drop in accuracy and other metrics when the calibration set does not cover all labels.
  One way to bypass it is to require substantially more calibration data. However, it can be impractical, as even five calibration examples per class would mean 50,000 (\(K = 10,000\)) additional labeled images for post-hoc calibration.  
  In contrast, our approaches are robust even when many classes are absent from the calibration set.
  The improvement is also evident in the cumulative mass calibration curves (see Figure~\ref{fig:cum_mass_two_datasets}), where our \emph{adaptive} TS with MSP nonconformity score and global quantile for  \(\alpha=0.1\) is nearly indistinguishable from the ideal curve, while also performing strongly on other metrics.


\section{Limitations}
\label{sec:limitations}
  There are several limitations of our approach.

  \begin{enumerate}
    \item There is a need to select a hyperparameter \(\alpha\) to apply our proposed algorithms. While one may use the same \(\alpha\) as for the plain CP procedure (there, it is dictated by application needs), in general, one may choose \(\alpha\) for calibration based on other considerations.
    One future direction could be to develop a procedure that automatically selects a specific \(\alpha\), or does not need it at all, while maintaining (marginal) CP guarantees.
\\
    \item Our adaptive TS calibration procedure uses classical \emph{marginally} valid conformal sets. However, as shown in the thoughtful procedure in Section~\ref{sec:background}, only valid \emph{conditional} coverage sets may lead to the full calibration.
  \end{enumerate}


\section{Conclusion}
\label{sec:conclusion}
  In this paper, we introduced two notions of \emph{cumulative mass calibration} that require calibration of cumulative probabilities, which is a necessary condition for full multiclass calibration. 
We defined \( \alpha \)-CMCE and CMCE, practical empirical measures of cumulative mass calibration.

Building on split CP, we propose an \emph{adaptive} TS calibration procedure that ensures \(\alpha\)-cumulative mass calibration.
  The proposed method keeps a finite-sample, distribution-free \emph{marginal} coverage guarantee at level \(1 - \alpha\) from CP. 
The calibrator is especially effective in settings with a large number of classes, consistently reducing \(\alpha\)-CMCE, CMCE, and other calibration metrics, despite not being explicitly optimized for it.

\bibliography{uai2026-template}

\newpage

\onecolumn

\title{Adaptive Cumulative Mass Calibration with Conformal Prediction\\
(Supplementary Material)}
\maketitle


\appendix

\section{Auxiliary theorems}
\label{sec:appendix_definition_check}

\begin{lemma}
For any \(\alpha \in [0,1]\), the ground-truth classifier \(\ptrue\) is \(\alpha\)-cumulative mass calibrated, meaning
\begin{equation*}
  \Pr\bigl(Y \in \HPRa[\ptrue(X)]\bigr) \ge 1 - \alpha.
\end{equation*}
\end{lemma}

\begin{proof}
Fix any \(x \in \XC\) and let \(q \in \Delta^K\) denote the conditional distribution \(q_k = \Pr(Y=k \mid X=x)\), so \(q = \ptrue(x)\).
Let \(S = \HPRa[q]\).
By definition of \(\HPRa[q]\), the set \(S\) contains the \(m\) most likely labels under \(q\), where \(m\) is the smallest integer such that:
\begin{equation*}
  \sum_{k = 1}^m q_{\pi(k)} \ge 1 - \alpha.
\end{equation*}
Since \(q_k = \Pr(Y=k \mid X=x)\), we have:
\begin{equation*}
  \Pr(Y \in S \mid X=x) = \sum_{k = 1}^m \Pr(Y=\pi(k) \mid X=x) = \sum_{k = 1}^m q_{\pi(k)}  \ge 1 - \alpha.
\end{equation*}

Now apply the law of total probability
\begin{equation*}
  \Pr\bigl(Y \in \HPRa[\ptrue(X)]\bigr)
  = \EE\Bigl[\indicator\bigl(Y \in \HPRa[\ptrue(X)] \mid X\bigr)\Bigr]
  \ge 1 - \alpha.
\end{equation*}
This proves the claim.
\end{proof}

\begin{remark}
Equality need not hold, even for \(\ptrue\).
By construction, \(\HPRa[q]\) is the smallest prefix of sorted probabilities whose cumulative sum reaches at least \(1-\alpha\).
If no prefix sum equals exactly \(1-\alpha\), then the last included class makes the cumulative mass strictly larger than \(1-\alpha\).
\end{remark}


\begin{lemma}
\label{lem:true_mass_calibration}
The ground-truth classifier \(\ptrue\) is cumulative mass calibrated.
\end{lemma}

\begin{proof}
By Definition~\ref{def:mass_calibration}, it suffices to show that \(\ptrue\) is \(\alpha\)-cumulative mass calibrated for an arbitrary \(\alpha \in [0,1]\).
Fix \(\alpha\).
For any \(x \in \XC\), let \(q = \ptrue(x)\) and let \(S_\alpha(x) = \HPRa[q]\).
By construction of \(\HPRa[q]\), we have
\begin{equation*}
  \sum_{k \in S_\alpha(x)} q_k \ge 1 - \alpha.
\end{equation*}
Since \(q_k = \Pr(Y=k \mid X=x)\), it follows that
\begin{equation*}
  \Pr\bigl(Y \in S_\alpha(x) \mid X=x\bigr)
  = \sum_{k \in S_\alpha(x)} \Pr(Y=k \mid X=x)
  = \sum_{k \in S_\alpha(x)} q_k
  \ge 1 - \alpha.
\end{equation*}
Taking expectation over \(X\) yields
\begin{equation*}
  \Pr\bigl(Y \in \HPRa[\ptrue(X)]\bigr)
  = \EE\Bigl[\indicator\bigl(Y \in \HPRa[\ptrue(X)] \mid X\bigr)\Bigr]
  \ge 1 - \alpha.
\end{equation*}
Since \(\alpha\) was arbitrary, \(\ptrue\) is \(\alpha\)-cumulative mass calibrated for every \(\alpha \in [0,1]\), and therefore it is cumulative mass calibrated.
\end{proof}

\section{Datasets Description}
\label{sec:appendix_datasets_description}
  We conduct experiments on four widely used image classification benchmarks:
  \begin{itemize}
    \item \textbf{CIFAR-10}~\citep{krizhevsky2009learning}: Consists of natural images of size $32 \times 32$ pixels, evenly distributed across 10 classes.

    \item \textbf{CIFAR-100}~\citep{krizhevsky2009learning}: Similar in structure to CIFAR-10, but with 100 classes.

    \item \textbf{ImageNet}~\citep{imagenet}: ImageNet is a large-scale hierarchical dataset of millions of labeled real-world images across thousands of object categories, widely used as a benchmark for computer vision research. We use a subset of ImageNet, comprising 34,745 images and 1,000 classes.

    \item \textbf{iNaturalist 2021}~\citep{vanhorn2021benchmarkingrepresentationlearningnatural}: The dataset consists of real-world photographs collected from the citizen-science platform iNaturalist. Images capture plants, animals, fungi, and other natural species in their natural environments. This dataset comprises 10,000 species. 
  \end{itemize}

\section{Naive CMCE algorithm}
\label{sec:appendix_naive_cmce}

In this section, we describe one of the algorithms proposed in this paper, the \emph{Naive CMCE} procedure.

Cumulative mass calibration (Definition~\ref{def:mass_calibration}) is a necessary condition for full multiclass calibration.
It is also stronger than an \(\alpha\)-cumulative mass calibration, since the latter targets only a single level \(\alpha\).
Our \emph{adaptive} temperature scaling algorithm cannot be tuned for all \(\alpha\) simultaneously.
It relies on conformal prediction, which constructs predictive sets \(\CC_\alpha(x)\) for a fixed, user-chosen \(\alpha\).

This motivates a simple baseline that directly calibrates a base classifier by reducing empirical CMCE on a hold-out dataset.
We refer to this baseline as \emph{Naive CMCE}.

We use temperature scaling, but we tune it differently from both adaptive and classical temperature scaling~\citep{guo2017calibration}.
Classical temperature scaling fits a single global temperature by maximizing predictive likelihood.
In Naive CMCE, we fit a single global temperature by minimizing empirical CMCE.

Let \(\pmodel(x) \in \Delta^K\) denote the predictive distribution of the pretrained model.
For \(\tau>0\), we define the temperature-scaled probabilities as
\begin{equation*}
  \pmodel_{\tau}(x) = \text{Softmax}\bigl[\log \pmodel(x) / \tau\bigr].
\end{equation*}
This transformation preserves the class ordering for each input \(x\) and adjusts only the confidence.

Given a hold-out calibration set \(\DC\), we select the temperature by
\begin{equation*}
  \hat{\tau} \in \arg\min_{\tau \in \TC} \mathrm{CMCE}\bigl(\pmodel_{\tau}; \DC\bigr),
\end{equation*}
where \(\TC \subset (0,\infty)\) is a predefined grid.
We use a grid search since CMCE is not differentiable with respect to \(\tau\).

We construct \(\TC\) to be uniform in logarithmic space.
We set \(\tau_{\min}=10^{-3}\) and \(\tau_{\max}=10^{3}\).

This baseline provides no distribution-free guarantees, including no coverage guarantees.
Empirically, however, minimizing CMCE often improves other calibration metrics as well.
But as our results show, minimizing CMCE usually improves other calibration metrics as well, supporting the importance of cumulative mass calibration.

\section{Models Description}
\label{sec:appendix_models_description}

\paragraph{Models.}
  We employ pretrained models available through the HuggingFace library and the Birder project.  
  \begin{itemize}
    \item \textbf{CIFAR-10.} We run experiments using ResNet-20 and ResNet-56~\citep{he2016deep} architectures pretrained on the dataset. 
    \item \textbf{CIFAR-100.} We use ResNet-56, as well as a Vision Transformer (ViT)~\citep{dosovitskiy2020image} with approximately 86M parameters, pretrained on ImageNet by Google.  
    \item \textbf{ImageNet.} We evaluate a Vision Transformer (ViT) with approximately 86M parameters pretrained on ImageNet.  
    \item \textbf{iNaturalist21.} We use a Vision Transformer (ViT) with approximately 91.6M parameters.  
  \end{itemize}

\section{Computational Constraints}
\label{sec:appendix_computational_constraits}

    \paragraph{Baselines omitted due to computational constraints.}
    Due to computational constraints, we do not evaluate all baselines on every dataset.
    In particular, Dirichlet calibration scales poorly with the number of classes.
    In the official implementation we used, fitting materializes a per-sample outer product
    $XX^{\top}$ during reparameterization, resulting in a tensor of shape
    $(n_{\text{cal}},\, C,\, C)$, where $n_{\text{cal}}$ is the number of calibration samples
    and $C$ is the number of classes.
    For ImageNet ($C{=}1000$) with $n_{\text{cal}}{=}6000$, this requires storing
    $6 \times 10^{9}$ floating-point values, and for iNaturalist the requirement exceeds
    $10^{12}$ values, which is not feasible in our experimental setup.
    
    \paragraph{Venn--Abers with incomplete class coverage.}
    We also omit Venn--Abers calibration when the calibration split does not include all classes.
    Venn--Abers is a binary calibrator and is commonly extended to the multiclass case via a
    one-vs-all reduction.
    This reduction requires, for each class, at least one positive example in the calibration data.
    On iNaturalist, the calibration split does not cover all classes; hence, some one-vs-all
    subproblems have no positive instances and the corresponding calibrations are undefined.
    As a result, the method cannot be applied reliably in this setting.

\section{Experimental Setup Details}
\label{sec:appendix_setup_details}

\paragraph{Pipeline.}
  Our evaluation pipeline is organized as follows:
  \begin{enumerate}
    \item We compute predictions (logits) with pretrained models for the datasets discussed above.

    \item We split the data into two subsets: \textit{calibration} and \textit{test}, according to the declared proportion. 
    We use 10 splits for computing metric distributions, and we allocate 20\% of the data for the calibration set.

    \item We train the calibrators on the calibration data. Hence, the methods receive logits and true classes.

    \item We compute the calibrators' prediction for the test part of the data and evaluate metrics.
  \end{enumerate}

\paragraph{Visualization.}
  \begin{itemize}
    \item \textbf{Result tables.}
      The tables report the mean and variance of each evaluation metric across different experiments.  
      Values highlighted in \textcolor{ForestGreen}{green} indicate improvements over the uncalibrated model, while values in \textcolor{red}{orange} denote degradations.  

      The best-performing results for each metric (except coverage) are emphasized with \textbf{\underline{bold and underlined}} formatting.  

    \item \textbf{Our methods arguments comparison table.} The table compares some metrics (usually CMCE) for different setups. \underline{Underlined} are these metrics, which show the best performance across columns (usually, it corresponds with the best choice of \(\alpha\)). With \textbf{bold} font are highlighted metrics, which outperform  across rows (usually, it corresponds with the best choice of our method and score type for a fixed \(\alpha\))

    \item \textbf{Cumulative mass calibration curve.} The graph curve illustrates the relationship between the cumulative predicted probability mass of top-ranked labels and the observed coverage of the truth class along one bin. A perfectly calibrated model corresponds to the diagonal line, while deviations from this line indicate over- or under-confidence. We use aggregation over 25 bins.
  \end{itemize}

\paragraph{Metrics.}
  We evaluate classifiers using the following metrics:
  \begin{itemize}
    \item \textbf{Brier Score} is the mean squared error between predicted probabilities and one-hot true labels:
      \begin{equation*}
        \text{BS} = \frac{1}{n}\sum_{i=1}^n \sum_{k=1}^K \bigl(\indicator\{y_i=k\} - \pmodel_{k}(x_i)\bigr)^2.
      \end{equation*}
    
    \item \textbf{NLL} (Negative Log-Likelihood) is the average log loss of the probability assigned to the true label:
      \begin{equation*}
        \text{NLL} = -\frac{1}{n}\sum_{i=1}^n \log \pmodel_{y_i}(x_i).
      \end{equation*}
    
    \item \textbf{ECE} (Expected Calibration Error)~\citep{guo2017calibration} is the average absolute calibration gap across \(b\) confidence bins:
      \begin{equation*}
        \text{ECE}=\sum_{i=1}^{b} \frac{|B_i|}{n} \bigl|\text{Acc}(B_i)-\text{Conf}(B_i)\bigr|,
      \end{equation*}
      where \(B_i\) is the \(i\)-th bin induced by predicted confidence, \(\text{Acc}(B_i)\) is the empirical accuracy in \(B_i\), and \(\text{Conf}(B_i)\) is the mean predicted confidence in \(B_i\).
    
    \item \textbf{cw-ECE} (Class-wise Expected Calibration Error)~\citep{kull2019beyond} averages per-class calibration gaps over \(b\) bins:
      \begin{equation*}
          \text{ECE}_{\text{cw}} = \sum_{j=1}^{K} \sum_{i=1}^{b} \frac{\bigl|B_i^{(j)}\bigr|}{n K} \cdot
          \bigl|\text{Acc}\bigl(B_i^{(j)}\bigr) - \text{Conf}\bigl(B_i^{(j)}\bigr)\bigr|,
      \end{equation*}
      where each bin \(B_i^{(j)}\) is obtained separately for each class.
    
    
  \end{itemize}


\section{Synthetic Experiment Details}
\label{sec:appendix_synthetic_experiment}

\begin{figure}
    \centering
    \includegraphics[width=0.5\linewidth]{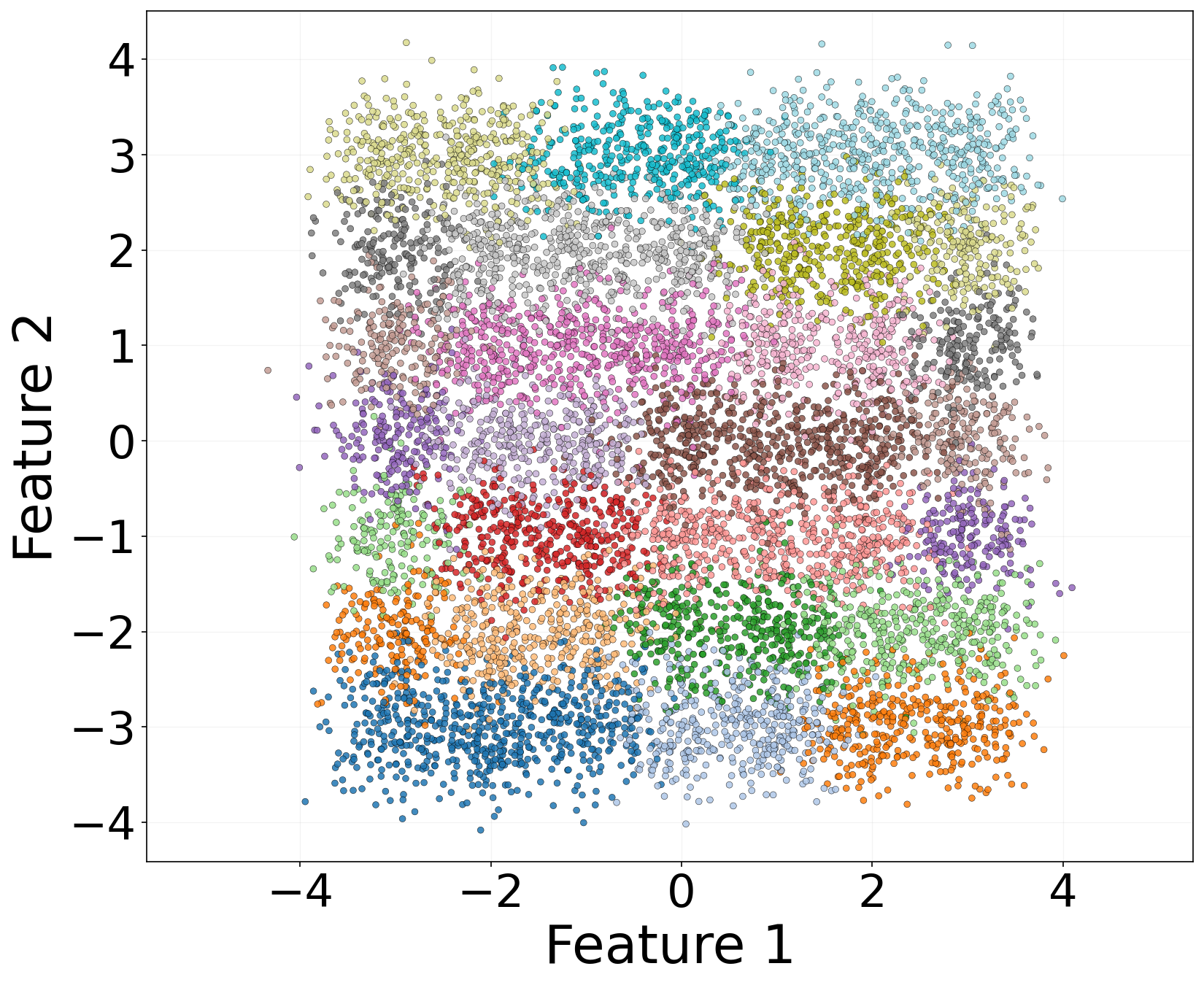}
    \caption{Samples from the synthetic dataset with \(K=49\); points are colored by class.}
    \label{fig:synthetic_experiment}
\end{figure}

\paragraph{Data.}
  In this synthetic experiment, we generate a 2D dataset with \(K\) classes, where \(K \) can be up to 10,000. 
  See example of a synthetic dataset in Figure~\ref{fig:synthetic_experiment}

  Each class \(k\) is represented by an isotropic Gaussian, with shared variance \(\sigma=0.35\) and centers, uniformly placed in a square grid. We draw labels uniformly, and sample inputs from an isotropic Gaussian around their class center:
  \begin{equation*}
    p(x \mid y=k) = \NC(\mu_k, \sigma^2 I). 
  \end{equation*}
  Given any \(x\), the ground-truth class probabilities form a normalized Gaussian mixture with equal priors:
  \begin{equation*}
    p(y=k \mid x) \propto \exp \left(-\tfrac{1}{2\sigma^2} \|x-\mu_k\|_2^2\right).
  \end{equation*}
  This construction is simple, gives high accuracy when \(\sigma\) is small, and still contains aleatoric uncertainty due to overlap between nearby classes.

\paragraph{Model, loss, and optimization.}
  We train a small, fully connected classifier network that maps \(x \in \RR^2\) to \(K\) logits. 
  To help with many-class separation on grids, we add random Fourier features (RFF; \citealp{liu2020simple}) before the MLP:
  \begin{equation*}
    \phi(x) = \bigl[x, \sin(2\pi B^\top x), \cos(2\pi B^\top x)\bigr],
  \end{equation*}
  where \(B \in \RR^{2 \times F}\) has i.i.d. Gaussian entries. 

  We train the classifier with cross-entropy loss and AdamW optimizer.

\section{Baselines Description}
\label{sec:appendix_baseline_description}
  In our experiments, we compare several post-hoc calibration methods. When a method is binary by design, we apply it to each class in a one-vs-rest (OvR) way and then renormalize:
  \begin{equation}
    q_k(x) = g_k\bigl(z_k(x)\bigr)
    \quad\Rightarrow\quad
    \tilde p_k(x) = \frac{q_k(x)}{\sum_{j = 1}^K q_j(x)},
  \label{eq:probability_renormalization}
  \end{equation}
  where \(z_k = z_k(x)\) is the score (or logit) for class \(k\) and \(g_k\) is the binary calibrator.

\paragraph{Platt scaling~\citep{platt1999probabilistic} (binary, we use OvR for multiclass).}
  Fits a sigmoid to map scores to probabilities:
  \begin{equation*}
    \widehat{p}(y_k = 1 \mid z_k) = \sigma(Az_k + B) =\frac{1}{1 + e^{-(A z_k + B)}}.
  \end{equation*}
  Then, we renormalize these binary probabilities as in~\eqref{eq:probability_renormalization} to obtain categorical probabilities.

\paragraph{Temperature scaling~\citep{guo2017calibration} (multiclass).}
  Divides all logits by a single positive scalar \(T > 0\) before softmax:
  \begin{equation*}
    \widehat{p}_i = \softmax \big(z / T \big)_i = \frac{e^{z_i/T}}{\sum_j e^{z_j/T}}.
  \end{equation*}
  This parameter is tuned via the maximization (gradient ascent) of the likelihood over the calibration split of the data.

\paragraph{Isotonic regression~\citep{zadrozny2002transforming} (binary, we use OvR for multiclass).}
  Learns a monotone, piecewise-constant map \(g\) from score to probability with no parametric form:
  \begin{equation*}
    \widehat{p}(y_k=1 \mid z_k) = g(z_k),
  \end{equation*}
  with \(g\) non-decreasing (which is achieved by the Pool Adjacent Violators Algorithm (PAVA)). Again, for categorical probabilities, we renormalize these binary probabilities as in~\eqref{eq:probability_renormalization}.

\paragraph{Dirichlet calibration~\citep{kull2019beyond} (multiclass).}
  Learns a parametric mapping on the simplex using features of the uncalibrated scores, implemented as a multinomial logistic regression:
  \begin{equation*}
    \widehat{p} = \softmax \bigl(W \phi(z) + b\bigr),
  \end{equation*}
  where \(\phi(z)\) are fixed transformations of scores (e.g., logits or log-softmax and simple interactions).  

\paragraph{Venn-Abers~\citep{vovk2014vennaberspredictors,vovk2015large} (binary; OvR for multiclass).}
  Venn-Abers turns a scoring binary classifier into a (multi-)probabilistic predictor with calibration guarantees: for a test input \(x\), it considers the two hypothetical labels \(y \in \{0, 1\}\), assigns \((x,y)\) to a Venn category induced by the score, and uses isotonic regression to produce a probability pair \((p_0, p_1)\) that act as a lower and upper estimate of true class probability.

  We report a single probability via the standard Venn-Abers aggregation; for multiclass classification, we use the one-vs-rest approach. We follow the IVAP implementation from~\citep{vovk2015large} (Algorithms 5-6) and utilize the publicly available code\footnote{\url{https://github.com/ip200/venn-abers/tree/main}}.
  In our experiments, we will emphasize that we use OvA (one-vs-all), which corresponds to the OvR strategy.

\paragraph{Sample-dependent adaptive temperature~\citep{joy2023sample}.}
  The idea is to predict temperature values on a per-data-point basis
  to address the limitation of a single temperature parameter, which could reduce the individual confidence of the predictions, irrespective of whether the classification of a given input is correct
  or incorrect.
  For this, they fit a separate temperature prediction module
  (a small neural network) that operates independently on each input sample.

  The calibrated probability is then computed as follows:
  \begin{equation*}
    \widehat{p}_i(x) = \softmax \bigl(z / T(x) \bigr)_i = \frac{e^{z_i/T(x)}}{\sum_j e^{z_j/T(x)}},
  \end{equation*}
  where \(z\) also depends on \(x\), but we omitted it for clarity.

\paragraph{Optimization and splits.}
  All calibrators are fit on a held-out calibration split only (test split is never used for fitting).  

\section{Additional Experimental Results}
\label{sec:appendix_additional_experimental_results}
  In this section, we present additional experimental results on the considered datasets. 
  In the tables we present below, we will use the following shortcuts to display the strategy for threshold selection (see Section~\ref{sec:construction_of_predictive_sets} in the main part): \texttt{C} stands for \emph{Constant} (identical transformation of the conformal threshold), and \texttt{Ada} stands for \emph{Adaptive}, as a particular way to incorporate dependency on \(x\) via a normalizing flow~\citep{colombo2024normalizing}.

\subsection{CIFAR-10}
  We consider a classification problem over CIFAR-10. We present complete results in Table~\ref{tab:results_cifar10_resnet20} (for ResNet20) and in Table~\ref{tab:results_cifar10_resnet56} (for ResNet56). Additionally, we present the best instances of our algorithms in Table~\ref{tab:results_cifar10_resnet20_best} and Table~\ref{tab:results_cifar10_resnet56_best} in a similar manner.
  We see that for such a small dataset, our algorithm does not provide a significant gain. 
  Nonetheless, our methods consistently outperform the uncalibrated model across all metrics. 
  In particular, Naive CMCE achieves the lowest CMCE while also improving NLL and Brier score compared to the base model. Temperature Scaling and Isotonic Regression remain strong baselines.

    The accuracy for ResNet20 and ResNet56 are same for all calibration methods, and are (\( 0.9256 \pm 0.0011 \)) and (\( 0.9439 \pm 0.001 \)) correspondingly.

  \begin{table*}[htbp]
\centering
\small
\caption{Evaluation results on CIFAR-10 and ResNet-20 (MSP score, Constant transformation, $\alpha=0.01$ for ours).
\textcolor{red}{Orange}/\textcolor{ForestGreen}{Green} indicate worse/better performance than the uncalibrated model. \textbf{Bold} and \underline{underline} mark the best and second-best values.}
\label{tab:results_cifar10_resnet20_best}
\resizebox{\textwidth}{!}{%
\begin{tabular}{@{}l @{\hspace{8pt}}c c c c c c c}
\toprule
Calibrator & ECE & MCE & cw-ECE & NLL & Brier Score & CMCE & $\alpha$-CMCE (0.01) \\
\midrule
Base
& $.04 \spm .00$ & $.32 \spm .21$ & $.01 \spm .00$ & $.282 \spm .005$ & $.12 \spm .00$ & $.007 \spm .000$ & $.007 \spm .001$ \\
\midrule
Ada-TS
& \textcolor{ForestGreen}{$.03 \spm .00$} & \textcolor{red}{$.35 \spm .23$} & \textcolor{ForestGreen}{$.01 \spm .00$} & \textcolor{red}{$.328 \spm .047$} & \textcolor{ForestGreen}{$.12 \spm .00$} & \textcolor{ForestGreen}{$.004 \spm .001$} & \textcolor{ForestGreen}{$.004 \spm .004$} \\
Dirichlet
& \textcolor{ForestGreen}{$\mathbf{.01 \spm .00}$} & \textcolor{red}{$.45 \spm .31$} & \textcolor{ForestGreen}{$\mathbf{.01 \spm .00}$} & \textcolor{ForestGreen}{$.242 \spm .007$} & \textcolor{ForestGreen}{$.12 \spm .00$} & \textcolor{ForestGreen}{$\underline{.002 \spm .001}$} & \textcolor{ForestGreen}{$.005 \spm .001$} \\
Isotonic
& \textcolor{ForestGreen}{$.02 \spm .00$} & \textcolor{red}{$.34 \spm .25$} & \textcolor{ForestGreen}{$\underline{.01 \spm .00}$} & \textcolor{red}{$.299 \spm .030$} & \textcolor{ForestGreen}{$.11 \spm .00$} & \textcolor{ForestGreen}{$.003 \spm .001$} & \textcolor{ForestGreen}{$.004 \spm .001$} \\
Platt Scaling
& \textcolor{ForestGreen}{$.04 \spm .00$} & \textcolor{ForestGreen}{$.29 \spm .05$} & \textcolor{ForestGreen}{$.01 \spm .00$} & \textcolor{ForestGreen}{$.256 \spm .002$} & \textcolor{ForestGreen}{$.12 \spm .00$} & \textcolor{ForestGreen}{$.006 \spm .001$} & \textcolor{red}{$.008 \spm .000$} \\
Temp. scaling
& \textcolor{ForestGreen}{$.02 \spm .01$} & \textcolor{red}{$.35 \spm .24$} & \textcolor{ForestGreen}{$.01 \spm .00$} & \textcolor{ForestGreen}{$.242 \spm .013$} & \textcolor{ForestGreen}{$.11 \spm .00$} & \textcolor{ForestGreen}{$.003 \spm .002$} & \textcolor{ForestGreen}{$\underline{.003 \spm .004}$} \\
V.-Abers (OvA)
& \textcolor{ForestGreen}{$.02 \spm .00$} & \textcolor{ForestGreen}{$\underline{.20 \spm .04}$} & \textcolor{ForestGreen}{$.01 \spm .00$} & \textcolor{ForestGreen}{$.246 \spm .003$} & \textcolor{ForestGreen}{$.11 \spm .00$} & \textcolor{red}{$.010 \spm .000$} & \textcolor{red}{$.009 \spm .001$} \\
\midrule
Naive CMCE (ours)
& \textcolor{ForestGreen}{$\underline{.01 \spm .00}$} & \textcolor{ForestGreen}{$\mathbf{.17 \spm .06}$} & \textcolor{ForestGreen}{$.01 \spm .00$} & \textcolor{ForestGreen}{$\mathbf{.235 \spm .004}$} & \textcolor{ForestGreen}{$\mathbf{.11 \spm .00}$} & \textcolor{ForestGreen}{$\mathbf{.002 \spm .000}$} & \textcolor{ForestGreen}{$.006 \spm .001$} \\
TS (ours)
& \textcolor{ForestGreen}{$.02 \spm .00$} & \textcolor{ForestGreen}{$.27 \spm .13$} & \textcolor{ForestGreen}{$.01 \spm .00$} & \textcolor{ForestGreen}{$\underline{.241 \spm .004}$} & \textcolor{ForestGreen}{$\underline{.11 \spm .00}$} & \textcolor{ForestGreen}{$.003 \spm .000$} & \textcolor{ForestGreen}{$\mathbf{.001 \spm .001}$} \\
\bottomrule
\end{tabular}
}
\end{table*}
  \begin{table*}[!ht]
\centering
\small
\caption{Evaluation results on CIFAR-10 and ResNet-56 (MSP score, Constant transformation, $\alpha=0.01$ for ours).
Metrics\textsuperscript{*} are scaled by \(100\). \textcolor{red}{Orange}/\textcolor{ForestGreen}{Green} indicate worse/better performance than the uncalibrated model. \textbf{Bold} and \underline{underline} mark the best and second-best values.}
\label{tab:results_cifar10_resnet56_best}
\resizebox{\textwidth}{!}{%
\begin{tabular}{@{}l @{\hspace{8pt}}c c c c c c c}
\toprule
Calibrator & ECE & MCE & cw-ECE & NLL & Brier Score & CMCE & $\alpha$-CMCE (0.01) \\
\midrule
Base
& $.04 \spm .00$ & $.41 \spm .03$ & $.01 \spm .00$ & $.254 \spm .006$ & $.09 \spm .00$ & $.006 \spm .000$ & $.010 \spm .001$ \\
\midrule
Ada-TS
& \textcolor{ForestGreen}{$.02 \spm .00$} & \textcolor{ForestGreen}{$.26 \spm .03$} & \textcolor{ForestGreen}{$.01 \spm .00$} & \textcolor{ForestGreen}{$.241 \spm .018$} & \textcolor{ForestGreen}{$.09 \spm .00$} & \textcolor{ForestGreen}{$\underline{.003 \spm .001}$} & \textcolor{ForestGreen}{$\mathbf{.000 \spm .002}$} \\
Dirichlet
& \textcolor{ForestGreen}{$.02 \spm .00$} & \textcolor{ForestGreen}{$.40 \spm .20$} & \textcolor{ForestGreen}{$\underline{.01 \spm .00}$} & \textcolor{ForestGreen}{$.202 \spm .007$} & \textcolor{ForestGreen}{$.09 \spm .00$} & \textcolor{ForestGreen}{$.003 \spm .001$} & \textcolor{ForestGreen}{$.004 \spm .001$} \\
Isotonic
& \textcolor{ForestGreen}{$\underline{.01 \spm .00}$} & \textcolor{ForestGreen}{$\mathbf{.22 \spm .02}$} & \textcolor{ForestGreen}{$\mathbf{.01 \spm .00}$} & \textcolor{red}{$.265 \spm .015$} & \textcolor{ForestGreen}{$\underline{.09 \spm .00}$} & \textcolor{ForestGreen}{$.003 \spm .000$} & \textcolor{ForestGreen}{$.003 \spm .000$} \\
Platt Scaling
& \textcolor{ForestGreen}{$.02 \spm .00$} & \textcolor{ForestGreen}{$.27 \spm .11$} & \textcolor{ForestGreen}{$.01 \spm .00$} & \textcolor{ForestGreen}{$\underline{.194 \spm .004}$} & \textcolor{ForestGreen}{$.09 \spm .00$} & \textcolor{ForestGreen}{$.004 \spm .001$} & \textcolor{ForestGreen}{$.006 \spm .001$} \\
Temp. scaling
& \textcolor{ForestGreen}{$\mathbf{.01 \spm .00}$} & \textcolor{ForestGreen}{$.25 \spm .03$} & \textcolor{ForestGreen}{$.01 \spm .00$} & \textcolor{ForestGreen}{$\mathbf{.193 \spm .004}$} & \textcolor{ForestGreen}{$\mathbf{.09 \spm .00}$} & \textcolor{ForestGreen}{$.003 \spm .000$} & \textcolor{ForestGreen}{$.006 \spm .001$} \\
V.-Abers (OvA)
& \textcolor{ForestGreen}{$.02 \spm .00$} & \textcolor{ForestGreen}{$\underline{.22 \spm .04}$} & \textcolor{ForestGreen}{$.01 \spm .00$} & \textcolor{ForestGreen}{$.200 \spm .002$} & \textcolor{ForestGreen}{$.09 \spm .00$} & \textcolor{red}{$.009 \spm .001$} & \textcolor{ForestGreen}{$.008 \spm .000$} \\
\midrule
Naive CMCE (ours)
& \textcolor{ForestGreen}{$.02 \spm .00$} & \textcolor{ForestGreen}{$.28 \spm .04$} & \textcolor{ForestGreen}{$.01 \spm .00$} & \textcolor{ForestGreen}{$.196 \spm .005$} & \textcolor{ForestGreen}{$.09 \spm .00$} & \textcolor{ForestGreen}{$\mathbf{.002 \spm .000}$} & \textcolor{ForestGreen}{$.004 \spm .001$} \\
TS (ours)
& \textcolor{ForestGreen}{$.02 \spm .00$} & \textcolor{ForestGreen}{$.34 \spm .05$} & \textcolor{ForestGreen}{$.01 \spm .00$} & \textcolor{ForestGreen}{$.200 \spm .005$} & \textcolor{ForestGreen}{$.09 \spm .00$} & \textcolor{ForestGreen}{$.003 \spm .000$} & \textcolor{ForestGreen}{$\underline{.001 \spm .001}$} \\
\bottomrule
\end{tabular}
}
\end{table*}
  \begin{table*}[htbp]
    \small
    \caption{Evaluation results on CIFAR-10 and ResNet-20. \textcolor{red}{Orange}/\textcolor{ForestGreen}{Green} indicate worse/better performance than the uncalibrated model. \textbf{Bold} and \underline{underline} mark the best and second-best values.}
    \label{tab:results_cifar10_resnet20}
    \resizebox{\textwidth}{!}{%
    \begin{tabular}{@{}l @{\hspace{2pt}}l @{\hspace{2pt}}l @{\hspace{2pt}}l @{\hspace{8pt}}c c c c c c c c c c}
    \toprule
    Calibrator & $\alpha$ & Score & Transf. & ECE & MCE & cw-ECE & NLL & Brier Score & CMCE & $\alpha$-CMCE (0.2) & $\alpha$-CMCE (0.1) & $\alpha$-CMCE (0.05) & $\alpha$-CMCE (0.01) \\
    \midrule
    Base &  &  &  & $.04 \spm .00$ & $.32 \spm .21$ & $.01 \spm .00$ & $.282 \spm .005$ & $.12 \spm .00$ & $.007 \spm .000$ & $.149 \spm .001$ & $.059 \spm .001$ & $.019 \spm .001$ & $.007 \spm .001$ \\
    \midrule
    Ada-TS &  &  &  & \textcolor{ForestGreen}{$.03 \spm .00$} & \textcolor{red}{$.35 \spm .23$} & \textcolor{ForestGreen}{$.01 \spm .00$} & \textcolor{red}{$.328 \spm .047$} & \textcolor{ForestGreen}{$.12 \spm .00$} & \textcolor{ForestGreen}{$.004 \spm .001$} & \textcolor{red}{$.156 \spm .004$} & \textcolor{red}{$.068 \spm .004$} & \textcolor{red}{$.026 \spm .005$} & \textcolor{ForestGreen}{$.004 \spm .004$} \\
    Dirichlet &  &  &  & \textcolor{ForestGreen}{$\mathbf{.01 \spm .00}$} & \textcolor{red}{$.45 \spm .31$} & \textcolor{ForestGreen}{$\mathbf{.01 \spm .00}$} & \textcolor{ForestGreen}{$.242 \spm .007$} & \textcolor{ForestGreen}{$.12 \spm .00$} & \textcolor{ForestGreen}{$\underline{.002 \spm .001}$} & \textcolor{red}{$.162 \spm .003$} & \textcolor{red}{$.076 \spm .001$} & \textcolor{red}{$.035 \spm .001$} & \textcolor{ForestGreen}{$.005 \spm .001$} \\
    Isotonic &  &  &  & \textcolor{ForestGreen}{$.02 \spm .00$} & \textcolor{red}{$.34 \spm .25$} & \textcolor{ForestGreen}{$\underline{.01 \spm .00}$} & \textcolor{red}{$.299 \spm .030$} & \textcolor{ForestGreen}{$.11 \spm .00$} & \textcolor{ForestGreen}{$.003 \spm .001$} & \textcolor{red}{$.165 \spm .001$} & \textcolor{red}{$.077 \spm .002$} & \textcolor{red}{$.034 \spm .002$} & \textcolor{ForestGreen}{$.004 \spm .001$} \\
    Platt Scaling &  &  &  & \textcolor{ForestGreen}{$.04 \spm .00$} & \textcolor{ForestGreen}{$.29 \spm .05$} & \textcolor{ForestGreen}{$.01 \spm .00$} & \textcolor{ForestGreen}{$.256 \spm .002$} & \textcolor{ForestGreen}{$.12 \spm .00$} & \textcolor{ForestGreen}{$.006 \spm .001$} & \textcolor{red}{$.169 \spm .001$} & \textcolor{red}{$.084 \spm .001$} & \textcolor{red}{$.040 \spm .001$} & \textcolor{red}{$.008 \spm .000$} \\
    Temp. scaling &  &  &  & \textcolor{ForestGreen}{$.02 \spm .01$} & \textcolor{red}{$.35 \spm .24$} & \textcolor{ForestGreen}{$.01 \spm .00$} & \textcolor{ForestGreen}{$.242 \spm .013$} & \textcolor{ForestGreen}{$.11 \spm .00$} & \textcolor{ForestGreen}{$.003 \spm .002$} & \textcolor{red}{$.159 \spm .005$} & \textcolor{red}{$.072 \spm .006$} & \textcolor{red}{$.032 \spm .005$} & \textcolor{ForestGreen}{$.003 \spm .004$} \\
    V.-Abers (OvA) &  &  &  & \textcolor{ForestGreen}{$.02 \spm .00$} & \textcolor{ForestGreen}{$\underline{.20 \spm .04}$} & \textcolor{ForestGreen}{$.01 \spm .00$} & \textcolor{ForestGreen}{$.246 \spm .003$} & \textcolor{ForestGreen}{$.11 \spm .00$} & \textcolor{red}{$.010 \spm .000$} & \textcolor{red}{$.173 \spm .001$} & \textcolor{red}{$.085 \spm .001$} & \textcolor{red}{$.043 \spm .000$} & \textcolor{red}{$.009 \spm .001$} \\
    \midrule
    Naive CMCE &  &  &  & \textcolor{ForestGreen}{$.01 \spm .00$} & \textcolor{ForestGreen}{$\mathbf{.17 \spm .06}$} & \textcolor{ForestGreen}{$.01 \spm .00$} & \textcolor{ForestGreen}{$\mathbf{.235 \spm .004}$} & \textcolor{ForestGreen}{$\mathbf{.11 \spm .00}$} & \textcolor{ForestGreen}{$\mathbf{.002 \spm .000}$} & \textcolor{red}{$.164 \spm .001$} & \textcolor{red}{$.077 \spm .002$} & \textcolor{red}{$.036 \spm .001$} & \textcolor{ForestGreen}{$.006 \spm .001$} \\
    TS (ours) & 0.01 & APS & Const & \textcolor{red}{$.16 \spm .00$} & \textcolor{ForestGreen}{$.24 \spm .02$} & \textcolor{red}{$.03 \spm .00$} & \textcolor{red}{$.375 \spm .003$} & \textcolor{red}{$.14 \spm .00$} & \textcolor{red}{$.063 \spm .001$} & \textcolor{red}{$.191 \spm .001$} & \textcolor{red}{$.098 \spm .000$} & \textcolor{red}{$.049 \spm .000$} & \textcolor{red}{$.010 \spm .000$} \\
    TS (ours) & 0.01 & APS & Ada & \textcolor{red}{$.04 \spm .00$} & \textcolor{red}{$.52 \spm .24$} & \textcolor{red}{$.01 \spm .00$} & \textcolor{red}{$.306 \spm .012$} & \textcolor{red}{$.12 \spm .00$} & \textcolor{red}{$.007 \spm .000$} & \textcolor{ForestGreen}{$.146 \spm .002$} & \textcolor{ForestGreen}{$.055 \spm .002$} & \textcolor{ForestGreen}{$.015 \spm .003$} & \textcolor{red}{$.012 \spm .002$} \\
    TS (ours) & 0.01 & MSP & Const & \textcolor{ForestGreen}{$.02 \spm .00$} & \textcolor{ForestGreen}{$.27 \spm .13$} & \textcolor{ForestGreen}{$.01 \spm .00$} & \textcolor{ForestGreen}{$\underline{.241 \spm .004}$} & \textcolor{ForestGreen}{$\underline{.11 \spm .00}$} & \textcolor{ForestGreen}{$.003 \spm .000$} & \textcolor{red}{$.160 \spm .001$} & \textcolor{red}{$.074 \spm .001$} & \textcolor{red}{$.034 \spm .001$} & \textcolor{ForestGreen}{$\underline{.001 \spm .001}$} \\
    TS (ours) & 0.01 & MSP & Ada & \textcolor{ForestGreen}{$.03 \spm .00$} & \textcolor{red}{$.42 \spm .27$} & \textcolor{ForestGreen}{$.01 \spm .00$} & \textcolor{ForestGreen}{$.272 \spm .013$} & \textcolor{ForestGreen}{$.12 \spm .00$} & \textcolor{ForestGreen}{$.004 \spm .001$} & \textcolor{red}{$.150 \spm .001$} & \textcolor{red}{$.061 \spm .003$} & \textcolor{red}{$.021 \spm .004$} & \textcolor{red}{$.008 \spm .004$} \\
    TS (ours) & 0.05 & APS & Const & \textcolor{red}{$.40 \spm .01$} & \textcolor{red}{$.62 \spm .00$} & \textcolor{red}{$.08 \spm .00$} & \textcolor{red}{$.801 \spm .010$} & \textcolor{red}{$.34 \spm .00$} & \textcolor{red}{$.186 \spm .003$} & \textcolor{red}{$.199 \spm .000$} & \textcolor{red}{$.100 \spm .000$} & \textcolor{red}{$.050 \spm .000$} & \textcolor{red}{$.010 \spm .000$} \\
    TS (ours) & 0.05 & APS & Ada & \textcolor{red}{$.05 \spm .00$} & \textcolor{red}{$.58 \spm .25$} & \textcolor{red}{$.01 \spm .00$} & \textcolor{red}{$.325 \spm .013$} & \textcolor{red}{$.13 \spm .00$} & \textcolor{red}{$.008 \spm .000$} & \textcolor{ForestGreen}{$\mathbf{.139 \spm .002}$} & \textcolor{ForestGreen}{$\underline{.046 \spm .002}$} & \textcolor{ForestGreen}{$\underline{.004 \spm .002}$} & \textcolor{red}{$.010 \spm .001$} \\
    TS (ours) & 0.05 & MSP & Const & \textcolor{ForestGreen}{$\underline{.01 \spm .00}$} & \textcolor{red}{$.55 \spm .19$} & \textcolor{ForestGreen}{$.01 \spm .00$} & \textcolor{red}{$.301 \spm .010$} & \textcolor{red}{$.12 \spm .00$} & \textcolor{red}{$.007 \spm .001$} & \textcolor{ForestGreen}{$.148 \spm .003$} & \textcolor{ForestGreen}{$.050 \spm .004$} & \textcolor{ForestGreen}{$\mathbf{.000 \spm .004}$} & \textcolor{ForestGreen}{$.002 \spm .001$} \\
    TS (ours) & 0.05 & MSP & Ada & \textcolor{ForestGreen}{$.03 \spm .01$} & \textcolor{red}{$.45 \spm .26$} & \textcolor{ForestGreen}{$.01 \spm .00$} & \textcolor{red}{$.301 \spm .015$} & \textcolor{red}{$.12 \spm .00$} & \textcolor{red}{$.009 \spm .001$} & \textcolor{ForestGreen}{$.146 \spm .003$} & \textcolor{ForestGreen}{$.053 \spm .004$} & \textcolor{ForestGreen}{$.010 \spm .005$} & \textcolor{ForestGreen}{$.002 \spm .001$} \\
    TS (ours) & 0.1 & APS & Const & \textcolor{red}{$.48 \spm .01$} & \textcolor{red}{$.77 \spm .00$} & \textcolor{red}{$.03 \spm .00$} & \textcolor{red}{$1.252 \spm .028$} & \textcolor{red}{$.48 \spm .01$} & \textcolor{red}{$.254 \spm .005$} & \textcolor{red}{$.199 \spm .000$} & \textcolor{red}{$.100 \spm .000$} & \textcolor{red}{$.050 \spm .000$} & \textcolor{red}{$.010 \spm .000$} \\
    TS (ours) & 0.1 & APS & Ada & \textcolor{red}{$.04 \spm .00$} & \textcolor{red}{$.76 \spm .19$} & \textcolor{red}{$.01 \spm .00$} & \textcolor{red}{$.328 \spm .013$} & \textcolor{red}{$.13 \spm .00$} & $.007 \spm .000$ & \textcolor{ForestGreen}{$\underline{.139 \spm .002}$} & \textcolor{ForestGreen}{$\mathbf{.044 \spm .002}$} & \textcolor{red}{$.021 \spm .002$} & \textcolor{red}{$.008 \spm .001$} \\
    TS (ours) & 0.1 & MSP & Const & \textcolor{red}{$.05 \spm .00$} & \textcolor{ForestGreen}{$.32 \spm .22$} & \textcolor{red}{$.01 \spm .00$} & \textcolor{red}{$.317 \spm .003$} & \textcolor{red}{$.13 \spm .00$} & \textcolor{red}{$.023 \spm .000$} & \textcolor{ForestGreen}{$.145 \spm .001$} & \textcolor{ForestGreen}{$.048 \spm .002$} & \textcolor{red}{$.037 \spm .001$} & \textcolor{ForestGreen}{$.005 \spm .001$} \\
    TS (ours) & 0.1 & MSP & Ada & \textcolor{red}{$.07 \spm .00$} & \textcolor{red}{$.43 \spm .25$} & \textcolor{red}{$.02 \spm .00$} & \textcolor{red}{$.326 \spm .005$} & \textcolor{red}{$.13 \spm .00$} & \textcolor{red}{$.023 \spm .000$} & \textcolor{ForestGreen}{$.146 \spm .002$} & \textcolor{ForestGreen}{$.053 \spm .002$} & \textcolor{red}{$.031 \spm .002$} & \textcolor{ForestGreen}{$.003 \spm .001$} \\
    TS (ours) & 0.2 & APS & Const & \textcolor{red}{$.25 \spm .00$} & \textcolor{red}{$.83 \spm .01$} & \textcolor{red}{$.03 \spm .00$} & \textcolor{red}{$.671 \spm .007$} & \textcolor{red}{$.28 \spm .00$} & \textcolor{red}{$.108 \spm .001$} & \textcolor{red}{$.159 \spm .001$} & \textcolor{red}{$.070 \spm .001$} & \textcolor{red}{$.029 \spm .001$} & \textcolor{ForestGreen}{$\mathbf{.000 \spm .001}$} \\
    TS (ours) & 0.2 & APS & Ada & \textcolor{red}{$.08 \spm .10$} & \textcolor{red}{$.64 \spm .26$} & \textcolor{red}{$.01 \spm .01$} & \textcolor{red}{$.381 \spm .172$} & \textcolor{red}{$.15 \spm .07$} & \textcolor{red}{$.026 \spm .046$} & \textcolor{ForestGreen}{$.141 \spm .008$} & \textcolor{red}{$.067 \spm .003$} & \textcolor{red}{$.024 \spm .003$} & \textcolor{ForestGreen}{$.005 \spm .003$} \\
    TS (ours) & 0.2 & MSP & Const & \textcolor{red}{$.17 \spm .00$} & $.32 \spm .21$ & \textcolor{red}{$.03 \spm .00$} & \textcolor{red}{$.406 \spm .005$} & \textcolor{red}{$.15 \spm .00$} & \textcolor{red}{$.055 \spm .001$} & $.149 \spm .001$ & \textcolor{red}{$.071 \spm .001$} & \textcolor{red}{$.032 \spm .001$} & \textcolor{ForestGreen}{$.004 \spm .001$} \\
    TS (ours) & 0.2 & MSP & Ada & \textcolor{red}{$.16 \spm .00$} & \textcolor{red}{$.34 \spm .20$} & \textcolor{red}{$.03 \spm .00$} & \textcolor{red}{$.401 \spm .002$} & \textcolor{red}{$.15 \spm .00$} & \textcolor{red}{$.054 \spm .001$} & \textcolor{ForestGreen}{$.146 \spm .003$} & \textcolor{red}{$.074 \spm .001$} & \textcolor{red}{$.033 \spm .001$} & \textcolor{ForestGreen}{$.003 \spm .001$} \\
    \bottomrule
    \end{tabular}
    }
    \end{table*}
  \begin{table*}[!t]
\small
\caption{Evaluation results on CIFAR-10 and ResNet-56.
\textcolor{red}{Orange}/\textcolor{ForestGreen}{Green} indicate worse/better performance than the uncalibrated model. \textbf{Bold} and \underline{underline} mark the best and second-best values.
}
\label{tab:results_cifar10_resnet56}
\resizebox{\textwidth}{!}{%
\begin{tabular}{@{}l @{\hspace{2pt}}l @{\hspace{2pt}}l @{\hspace{2pt}}l @{\hspace{8pt}}c c c c c c c c c c}
\toprule
Calibrator & $\alpha$ & Score & Transf. & ECE & MCE & cw-ECE & NLL & Brier Score & CMCE & $\alpha$-CMCE (0.2) & $\alpha$-CMCE (0.1) & $\alpha$-CMCE (0.05) & $\alpha$-CMCE (0.01) \\
\midrule
Base &  &  &  & $.04 \spm .00$ & $.41 \spm .03$ & $.01 \spm .00$ & $.254 \spm .006$ & $.09 \spm .00$ & $.006 \spm .000$ & $.157 \spm .001$ & $.064 \spm .001$ & $.020 \spm .001$ & $.010 \spm .001$ \\
\midrule
Ada-TS &  &  &  & \textcolor{ForestGreen}{$.02 \spm .00$} & \textcolor{ForestGreen}{$.26 \spm .03$} & \textcolor{ForestGreen}{$.01 \spm .00$} & \textcolor{ForestGreen}{$.241 \spm .018$} & \textcolor{ForestGreen}{$.09 \spm .00$} & \textcolor{ForestGreen}{$\underline{.003 \spm .001}$} & \textcolor{red}{$.169 \spm .002$} & \textcolor{red}{$.077 \spm .002$} & \textcolor{red}{$.034 \spm .002$} & \textcolor{ForestGreen}{$\mathbf{.000 \spm .002}$} \\
Dirichlet &  &  &  & \textcolor{ForestGreen}{$.02 \spm .00$} & \textcolor{ForestGreen}{$.40 \spm .20$} & \textcolor{ForestGreen}{$\underline{.01 \spm .00}$} & \textcolor{ForestGreen}{$.202 \spm .007$} & \textcolor{ForestGreen}{$.09 \spm .00$} & \textcolor{ForestGreen}{$.003 \spm .001$} & \textcolor{red}{$.166 \spm .002$} & \textcolor{red}{$.076 \spm .002$} & \textcolor{red}{$.033 \spm .002$} & \textcolor{ForestGreen}{$.004 \spm .001$} \\
Isotonic &  &  &  & \textcolor{ForestGreen}{$.01 \spm .00$} & \textcolor{ForestGreen}{$\mathbf{.22 \spm .02}$} & \textcolor{ForestGreen}{$\mathbf{.01 \spm .00}$} & \textcolor{red}{$.265 \spm .015$} & \textcolor{ForestGreen}{$\underline{.09 \spm .00}$} & \textcolor{ForestGreen}{$.003 \spm .000$} & \textcolor{red}{$.172 \spm .002$} & \textcolor{red}{$.080 \spm .002$} & \textcolor{red}{$.036 \spm .001$} & \textcolor{ForestGreen}{$.003 \spm .000$} \\
Platt Scaling &  &  &  & \textcolor{ForestGreen}{$.02 \spm .00$} & \textcolor{ForestGreen}{$.27 \spm .11$} & \textcolor{ForestGreen}{$.01 \spm .00$} & \textcolor{ForestGreen}{$\underline{.194 \spm .004}$} & \textcolor{ForestGreen}{$.09 \spm .00$} & \textcolor{ForestGreen}{$.004 \spm .001$} & \textcolor{red}{$.169 \spm .002$} & \textcolor{red}{$.079 \spm .002$} & \textcolor{red}{$.035 \spm .001$} & \textcolor{ForestGreen}{$.006 \spm .001$} \\
Temp. scaling &  &  &  & \textcolor{ForestGreen}{$\underline{.01 \spm .00}$} & \textcolor{ForestGreen}{$.25 \spm .03$} & \textcolor{ForestGreen}{$.01 \spm .00$} & \textcolor{ForestGreen}{$\mathbf{.193 \spm .004}$} & \textcolor{ForestGreen}{$\mathbf{.09 \spm .00}$} & \textcolor{ForestGreen}{$.003 \spm .000$} & \textcolor{red}{$.168 \spm .002$} & \textcolor{red}{$.079 \spm .001$} & \textcolor{red}{$.035 \spm .001$} & \textcolor{ForestGreen}{$.006 \spm .001$} \\
V.-Abers (OvA) &  &  &  & \textcolor{ForestGreen}{$.02 \spm .00$} & \textcolor{ForestGreen}{$\underline{.22 \spm .04}$} & \textcolor{ForestGreen}{$.01 \spm .00$} & \textcolor{ForestGreen}{$.200 \spm .002$} & \textcolor{ForestGreen}{$.09 \spm .00$} & \textcolor{red}{$.009 \spm .001$} & \textcolor{red}{$.179 \spm .001$} & \textcolor{red}{$.087 \spm .001$} & \textcolor{red}{$.042 \spm .001$} & \textcolor{ForestGreen}{$.008 \spm .000$} \\
\midrule
Naive CMCE &  &  &  & \textcolor{ForestGreen}{$.02 \spm .00$} & \textcolor{ForestGreen}{$.28 \spm .04$} & \textcolor{ForestGreen}{$.01 \spm .00$} & \textcolor{ForestGreen}{$.196 \spm .005$} & \textcolor{ForestGreen}{$.09 \spm .00$} & \textcolor{ForestGreen}{$\mathbf{.002 \spm .000}$} & \textcolor{red}{$.166 \spm .002$} & \textcolor{red}{$.075 \spm .002$} & \textcolor{red}{$.032 \spm .001$} & \textcolor{ForestGreen}{$.004 \spm .001$} \\
TS (ours) & 0.01 & APS & Const & \textcolor{red}{$.08 \spm .00$} & \textcolor{ForestGreen}{$.26 \spm .00$} & \textcolor{red}{$.02 \spm .00$} & \textcolor{ForestGreen}{$.253 \spm .006$} & \textcolor{ForestGreen}{$.09 \spm .00$} & \textcolor{red}{$.036 \spm .002$} & \textcolor{red}{$.183 \spm .001$} & \textcolor{red}{$.093 \spm .000$} & \textcolor{red}{$.047 \spm .000$} & \textcolor{ForestGreen}{$.010 \spm .000$} \\
TS (ours) & 0.01 & APS & Ada & \textcolor{red}{$.04 \spm .00$} & \textcolor{red}{$.45 \spm .08$} & \textcolor{red}{$.01 \spm .00$} & \textcolor{red}{$.279 \spm .014$} & \textcolor{red}{$.10 \spm .00$} & \textcolor{red}{$.007 \spm .000$} & \textcolor{ForestGreen}{$.154 \spm .002$} & \textcolor{ForestGreen}{$.060 \spm .002$} & \textcolor{ForestGreen}{$.015 \spm .003$} & \textcolor{red}{$.015 \spm .003$} \\
TS (ours) & 0.01 & MSP & Const & \textcolor{ForestGreen}{$.02 \spm .00$} & \textcolor{ForestGreen}{$.34 \spm .05$} & \textcolor{ForestGreen}{$.01 \spm .00$} & \textcolor{ForestGreen}{$.200 \spm .005$} & \textcolor{ForestGreen}{$.09 \spm .00$} & \textcolor{ForestGreen}{$.003 \spm .000$} & \textcolor{red}{$.163 \spm .001$} & \textcolor{red}{$.072 \spm .001$} & \textcolor{red}{$.030 \spm .001$} & \textcolor{ForestGreen}{$\underline{.001 \spm .001}$} \\
TS (ours) & 0.01 & MSP & Ada & \textcolor{ForestGreen}{$.03 \spm .00$} & \textcolor{ForestGreen}{$.40 \spm .05$} & \textcolor{ForestGreen}{$.01 \spm .00$} & \textcolor{ForestGreen}{$.240 \spm .019$} & \textcolor{ForestGreen}{$.09 \spm .00$} & \textcolor{ForestGreen}{$.004 \spm .001$} & \textcolor{red}{$.157 \spm .003$} & \textcolor{red}{$.065 \spm .004$} & \textcolor{red}{$.021 \spm .004$} & \textcolor{red}{$.010 \spm .003$} \\
TS (ours) & 0.05 & APS & Const & \textcolor{red}{$.22 \spm .01$} & \textcolor{red}{$.56 \spm .01$} & \textcolor{red}{$.05 \spm .00$} & \textcolor{red}{$.450 \spm .015$} & \textcolor{red}{$.17 \spm .01$} & \textcolor{red}{$.107 \spm .004$} & \textcolor{red}{$.195 \spm .000$} & \textcolor{red}{$.098 \spm .000$} & \textcolor{red}{$.049 \spm .000$} & $.010 \spm .000$ \\
TS (ours) & 0.05 & APS & Ada & \textcolor{red}{$.04 \spm .00$} & \textcolor{red}{$.52 \spm .18$} & \textcolor{red}{$.01 \spm .00$} & \textcolor{red}{$.289 \spm .008$} & \textcolor{red}{$.10 \spm .00$} & \textcolor{red}{$.007 \spm .000$} & \textcolor{ForestGreen}{$\mathbf{.150 \spm .002}$} & \textcolor{ForestGreen}{$.054 \spm .002$} & \textcolor{ForestGreen}{$\underline{.007 \spm .003}$} & \textcolor{red}{$.011 \spm .001$} \\
TS (ours) & 0.05 & MSP & Const & \textcolor{ForestGreen}{$\mathbf{.00 \spm .00}$} & \textcolor{red}{$.60 \spm .16$} & \textcolor{ForestGreen}{$.01 \spm .00$} & \textcolor{red}{$.282 \spm .019$} & \textcolor{red}{$.10 \spm .00$} & \textcolor{red}{$.010 \spm .001$} & \textcolor{ForestGreen}{$\mathbf{.150 \spm .003}$} & \textcolor{ForestGreen}{$\mathbf{.050 \spm .003}$} & \textcolor{ForestGreen}{$\mathbf{.000 \spm .003}$} & \textcolor{ForestGreen}{$.001 \spm .001$} \\
TS (ours) & 0.05 & MSP & Ada & \textcolor{ForestGreen}{$.03 \spm .01$} & \textcolor{red}{$.44 \spm .07$} & \textcolor{ForestGreen}{$.01 \spm .00$} & \textcolor{ForestGreen}{$.248 \spm .008$} & \textcolor{red}{$.10 \spm .00$} & \textcolor{red}{$.013 \spm .001$} & \textcolor{ForestGreen}{$.155 \spm .002$} & \textcolor{ForestGreen}{$.060 \spm .003$} & \textcolor{ForestGreen}{$.014 \spm .004$} & \textcolor{ForestGreen}{$.003 \spm .001$} \\
TS (ours) & 0.1 & APS & Const & \textcolor{red}{$.30 \spm .01$} & \textcolor{red}{$.70 \spm .04$} & \textcolor{red}{$.04 \spm .00$} & \textcolor{red}{$.677 \spm .016$} & \textcolor{red}{$.25 \spm .01$} & \textcolor{red}{$.155 \spm .004$} & \textcolor{red}{$.197 \spm .000$} & \textcolor{red}{$.099 \spm .000$} & \textcolor{red}{$.049 \spm .000$} & \textcolor{ForestGreen}{$.009 \spm .000$} \\
TS (ours) & 0.1 & APS & Ada & \textcolor{red}{$.04 \spm .00$} & \textcolor{red}{$.74 \spm .23$} & \textcolor{ForestGreen}{$.01 \spm .00$} & \textcolor{red}{$.271 \spm .010$} & \textcolor{red}{$.10 \spm .00$} & \textcolor{ForestGreen}{$.006 \spm .001$} & \textcolor{ForestGreen}{$\underline{.150 \spm .002}$} & \textcolor{ForestGreen}{$\underline{.054 \spm .002}$} & \textcolor{red}{$.024 \spm .003$} & \textcolor{ForestGreen}{$.009 \spm .003$} \\
TS (ours) & 0.1 & MSP & Const & \textcolor{red}{$.08 \spm .00$} & $.41 \spm .03$ & \textcolor{red}{$.02 \spm .00$} & \textcolor{red}{$.270 \spm .005$} & \textcolor{red}{$.10 \spm .00$} & \textcolor{red}{$.032 \spm .000$} & $.157 \spm .001$ & $.064 \spm .001$ & \textcolor{red}{$.039 \spm .001$} & \textcolor{ForestGreen}{$.007 \spm .000$} \\
TS (ours) & 0.1 & MSP & Ada & \textcolor{red}{$.08 \spm .00$} & \textcolor{ForestGreen}{$.38 \spm .07$} & \textcolor{red}{$.02 \spm .00$} & \textcolor{red}{$.283 \spm .005$} & \textcolor{red}{$.10 \spm .00$} & \textcolor{red}{$.030 \spm .001$} & \textcolor{ForestGreen}{$.155 \spm .002$} & \textcolor{ForestGreen}{$.060 \spm .003$} & \textcolor{red}{$.035 \spm .002$} & \textcolor{ForestGreen}{$.005 \spm .001$} \\
TS (ours) & 0.2 & APS & Const & \textcolor{red}{$.29 \spm .01$} & \textcolor{red}{$.81 \spm .00$} & \textcolor{red}{$.04 \spm .00$} & \textcolor{red}{$.645 \spm .014$} & \textcolor{red}{$.25 \spm .01$} & \textcolor{red}{$.133 \spm .003$} & \textcolor{red}{$.175 \spm .001$} & \textcolor{red}{$.082 \spm .001$} & \textcolor{red}{$.037 \spm .001$} & \textcolor{ForestGreen}{$.004 \spm .001$} \\
TS (ours) & 0.2 & APS & Ada & \textcolor{ForestGreen}{$.03 \spm .00$} & \textcolor{red}{$.77 \spm .05$} & \textcolor{ForestGreen}{$.01 \spm .00$} & \textcolor{red}{$.263 \spm .010$} & \textcolor{red}{$.10 \spm .00$} & \textcolor{red}{$.008 \spm .003$} & \textcolor{ForestGreen}{$.152 \spm .004$} & \textcolor{red}{$.075 \spm .005$} & \textcolor{red}{$.029 \spm .004$} & \textcolor{ForestGreen}{$.005 \spm .002$} \\
TS (ours) & 0.2 & MSP & Const & \textcolor{red}{$.18 \spm .00$} & \textcolor{ForestGreen}{$.39 \spm .00$} & \textcolor{red}{$.03 \spm .00$} & \textcolor{red}{$.388 \spm .006$} & \textcolor{red}{$.13 \spm .00$} & \textcolor{red}{$.064 \spm .001$} & $.157 \spm .001$ & \textcolor{red}{$.071 \spm .001$} & \textcolor{red}{$.027 \spm .001$} & \textcolor{ForestGreen}{$.003 \spm .001$} \\
TS (ours) & 0.2 & MSP & Ada & \textcolor{red}{$.18 \spm .00$} & \textcolor{ForestGreen}{$.39 \spm .00$} & \textcolor{red}{$.03 \spm .00$} & \textcolor{red}{$.383 \spm .004$} & \textcolor{red}{$.13 \spm .00$} & \textcolor{red}{$.063 \spm .000$} & $.157 \spm .001$ & \textcolor{red}{$.072 \spm .001$} & \textcolor{red}{$.029 \spm .001$} & \textcolor{ForestGreen}{$.002 \spm .001$} \\
\bottomrule
\end{tabular}
}
\end{table*}


\subsection{CIFAR-100}
  Here, we consider classification over CIFAR-100. 
  For this dataset, we consider ResNet-56 and Visual Transformer (ViT).
  Similarly to CIFAR-10, here we present full results in Table~\ref{tab:results_cifar100_vit} (ViT) and Table~\ref{tab:results_cifar100_resnet56} (ResNet56); in Tables~\ref{tab:results_cifar100_vit_best} and~\ref{tab:results_cifar100_inaturalist_best} (main text) we present the best results.
  
  We observe two patterns in our results: (i) performance is model-dependent, and calibrators tend to help ResNet-56 more than ViT; and (ii) Temperature Scaling often provides the best ECE/NLL among classical baselines, while our Naive CMCE yields the best (or near-best) CMCE and the tightest control of coverage near the target. 

    The accuracy for ResNet56 architecture are same ($0.7264 \pm 0.0023$) and for ViT  ($0.8947 \pm 0.0011$).

  
  \begin{table*}[htbp]
\small
\caption{Evaluation results on CIFAR-100 and ResNet-56. \textcolor{red}{Orange}/\textcolor{ForestGreen}{Green} indicate worse/better performance than the uncalibrated model. \textbf{Bold} and \underline{underline} mark the best and second-best values.}
\label{tab:results_cifar100_resnet56}
\resizebox{\textwidth}{!}{%
\begin{tabular}{@{}l @{\hspace{2pt}}l @{\hspace{2pt}}l @{\hspace{2pt}}l @{\hspace{8pt}}c c c c c c c c c c}
\toprule
Calibrator & $\alpha$ & Score & Transf. & ECE & MCE & cw-ECE & NLL & Brier Score & CMCE & $\alpha$-CMCE (0.2) & $\alpha$-CMCE (0.1) & $\alpha$-CMCE (0.05) & $\alpha$-CMCE (0.01) \\
\midrule
Base &  &  &  & $.14 \spm .00$ & $.32 \spm .02$ & $.00 \spm .00$ & $1.288 \spm .011$ & $.42 \spm .00$ & $.013 \spm .000$ & $.002 \spm .002$ & $.061 \spm .002$ & $.081 \spm .001$ & $.066 \spm .001$ \\
\midrule
Ada-TS &  &  &  & \textcolor{red}{$.15 \spm .00$} & \textcolor{ForestGreen}{$.29 \spm .03$} & \textcolor{ForestGreen}{$.00 \spm .00$} & \textcolor{red}{$2.368 \spm .047$} & \textcolor{red}{$.45 \spm .00$} & \textcolor{red}{$.019 \spm .001$} & \textcolor{red}{$.021 \spm .003$} & \textcolor{ForestGreen}{$.055 \spm .003$} & \textcolor{red}{$.087 \spm .003$} & \textcolor{red}{$.099 \spm .002$} \\
Dirichlet &  &  &  & \textcolor{red}{$.41 \spm .01$} & \textcolor{red}{$.74 \spm .04$} & \textcolor{red}{$.01 \spm .00$} & \textcolor{red}{$10.440 \spm .545$} & \textcolor{red}{$.82 \spm .02$} & \textcolor{red}{$.034 \spm .002$} & \textcolor{red}{$.212 \spm .009$} & \textcolor{red}{$.310 \spm .010$} & \textcolor{red}{$.358 \spm .011$} & \textcolor{red}{$.393 \spm .012$} \\
Isotonic &  &  &  & \textcolor{ForestGreen}{$.06 \spm .00$} & \textcolor{red}{$.33 \spm .34$} & \textcolor{ForestGreen}{$\underline{.00 \spm .00}$} & \textcolor{red}{$1.728 \spm .070$} & \textcolor{ForestGreen}{$.39 \spm .00$} & \textcolor{red}{$.017 \spm .002$} & \textcolor{red}{$.062 \spm .004$} & \textcolor{ForestGreen}{$\mathbf{.001 \spm .004}$} & \textcolor{ForestGreen}{$.023 \spm .003$} & \textcolor{ForestGreen}{$.031 \spm .004$} \\
Platt Scaling &  &  &  & \textcolor{red}{$.15 \spm .00$} & \textcolor{ForestGreen}{$.22 \spm .01$} & \textcolor{red}{$.00 \spm .00$} & \textcolor{ForestGreen}{$1.139 \spm .006$} & \textcolor{ForestGreen}{$.42 \spm .00$} & \textcolor{red}{$.015 \spm .001$} & \textcolor{red}{$.131 \spm .002$} & \textcolor{red}{$.068 \spm .001$} & \textcolor{ForestGreen}{$.034 \spm .001$} & \textcolor{ForestGreen}{$.007 \spm .000$} \\
Temp. scaling &  &  &  & \textcolor{ForestGreen}{$\mathbf{.03 \spm .00}$} & \textcolor{ForestGreen}{$\underline{.11 \spm .03}$} & \textcolor{ForestGreen}{$\mathbf{.00 \spm .00}$} & \textcolor{ForestGreen}{$\mathbf{1.038 \spm .007}$} & \textcolor{ForestGreen}{$\underline{.38 \spm .00}$} & \textcolor{ForestGreen}{$.004 \spm .000$} & \textcolor{red}{$.093 \spm .001$} & \textcolor{ForestGreen}{$.039 \spm .002$} & \textcolor{ForestGreen}{$.014 \spm .001$} & \textcolor{ForestGreen}{$.002 \spm .001$} \\
V.-Abers (OvA) &  &  &  & \textcolor{ForestGreen}{$.08 \spm .00$} & \textcolor{ForestGreen}{$.17 \spm .01$} & \textcolor{ForestGreen}{$.00 \spm .00$} & \textcolor{ForestGreen}{$1.119 \spm .008$} & \textcolor{ForestGreen}{$.39 \spm .00$} & \textcolor{red}{$.038 \spm .002$} & \textcolor{red}{$.138 \spm .003$} & \textcolor{red}{$.077 \spm .002$} & \textcolor{ForestGreen}{$.044 \spm .001$} & \textcolor{ForestGreen}{$.010 \spm .000$} \\
\midrule
Naive CMCE &  &  &  & \textcolor{ForestGreen}{$.04 \spm .00$} & \textcolor{ForestGreen}{$\mathbf{.11 \spm .01}$} & \textcolor{ForestGreen}{$\mathbf{.00 \spm .00}$} & \textcolor{ForestGreen}{$1.041 \spm .007$} & \textcolor{ForestGreen}{$.38 \spm .00$} & \textcolor{ForestGreen}{$\mathbf{.001 \spm .001}$} & \textcolor{red}{$.081 \spm .004$} & \textcolor{ForestGreen}{$.027 \spm .003$} & \textcolor{ForestGreen}{$\underline{.006 \spm .003}$} & \textcolor{ForestGreen}{$.003 \spm .002$} \\
TS (ours) & 0.01 & APS & Const & \textcolor{red}{$.18 \spm .01$} & \textcolor{red}{$.41 \spm .02$} & \textcolor{red}{$.00 \spm .00$} & \textcolor{red}{$1.417 \spm .025$} & \textcolor{red}{$.45 \spm .01$} & \textcolor{red}{$.098 \spm .005$} & \textcolor{red}{$.169 \spm .002$} & \textcolor{red}{$.087 \spm .002$} & \textcolor{ForestGreen}{$.045 \spm .001$} & \textcolor{ForestGreen}{$.009 \spm .000$} \\
TS (ours) & 0.01 & APS & Ada & \textcolor{red}{$.15 \spm .02$} & \textcolor{red}{$.35 \spm .04$} & \textcolor{red}{$.00 \spm .00$} & \textcolor{red}{$1.337 \spm .062$} & \textcolor{red}{$.43 \spm .02$} & \textcolor{red}{$.031 \spm .041$} & \textcolor{red}{$.034 \spm .077$} & \textcolor{ForestGreen}{$.033 \spm .067$} & \textcolor{ForestGreen}{$.057 \spm .057$} & \textcolor{ForestGreen}{$.052 \spm .035$} \\
TS (ours) & 0.01 & MSP & Const & \textcolor{ForestGreen}{$\underline{.04 \spm .00}$} & \textcolor{ForestGreen}{$.22 \spm .03$} & \textcolor{ForestGreen}{$\mathbf{.00 \spm .00}$} & \textcolor{ForestGreen}{$1.046 \spm .007$} & \textcolor{ForestGreen}{$.38 \spm .00$} & \textcolor{ForestGreen}{$.008 \spm .001$} & \textcolor{red}{$.092 \spm .003$} & \textcolor{ForestGreen}{$.035 \spm .002$} & \textcolor{ForestGreen}{$.014 \spm .002$} & \textcolor{ForestGreen}{$.002 \spm .001$} \\
TS (ours) & 0.01 & MSP & Ada & \textcolor{red}{$.14 \spm .00$} & \textcolor{red}{$.33 \spm .02$} & $.00 \spm .00$ & \textcolor{red}{$1.301 \spm .013$} & \textcolor{red}{$.42 \spm .00$} & $.013 \spm .000$ & \textcolor{ForestGreen}{$.001 \spm .002$} & \textcolor{red}{$.062 \spm .002$} & \textcolor{red}{$.083 \spm .002$} & \textcolor{red}{$.069 \spm .002$} \\
TS (ours) & 0.05 & APS & Const & \textcolor{red}{$.19 \spm .01$} & \textcolor{red}{$.39 \spm .01$} & \textcolor{red}{$.00 \spm .00$} & \textcolor{red}{$1.404 \spm .029$} & \textcolor{red}{$.45 \spm .01$} & \textcolor{red}{$.092 \spm .004$} & \textcolor{red}{$.164 \spm .001$} & \textcolor{red}{$.083 \spm .001$} & \textcolor{ForestGreen}{$.041 \spm .001$} & \textcolor{ForestGreen}{$.008 \spm .001$} \\
TS (ours) & 0.05 & APS & Ada & \textcolor{red}{$.15 \spm .00$} & \textcolor{red}{$.47 \spm .27$} & $.00 \spm .00$ & \textcolor{red}{$1.333 \spm .013$} & \textcolor{red}{$.42 \spm .00$} & $.013 \spm .000$ & \textcolor{red}{$.003 \spm .002$} & \textcolor{red}{$.067 \spm .003$} & \textcolor{red}{$.088 \spm .002$} & \textcolor{red}{$.069 \spm .001$} \\
TS (ours) & 0.05 & MSP & Const & \textcolor{ForestGreen}{$.04 \spm .00$} & \textcolor{ForestGreen}{$.12 \spm .01$} & \textcolor{ForestGreen}{$\mathbf{.00 \spm .00}$} & \textcolor{ForestGreen}{$\underline{1.039 \spm .008}$} & \textcolor{ForestGreen}{$\mathbf{.38 \spm .00}$} & \textcolor{ForestGreen}{$\underline{.004 \spm .000}$} & \textcolor{red}{$.074 \spm .003$} & \textcolor{ForestGreen}{$.022 \spm .002$} & \textcolor{ForestGreen}{$\mathbf{.003 \spm .002}$} & \textcolor{ForestGreen}{$\underline{.001 \spm .001}$} \\
TS (ours) & 0.05 & MSP & Ada & \textcolor{ForestGreen}{$.13 \spm .00$} & \textcolor{red}{$.33 \spm .03$} & \textcolor{ForestGreen}{$.00 \spm .00$} & \textcolor{red}{$1.315 \spm .010$} & \textcolor{red}{$.42 \spm .00$} & \textcolor{ForestGreen}{$.012 \spm .000$} & \textcolor{red}{$.002 \spm .001$} & \textcolor{red}{$.069 \spm .001$} & \textcolor{red}{$.093 \spm .001$} & \textcolor{ForestGreen}{$.061 \spm .002$} \\
TS (ours) & 0.1 & APS & Const & \textcolor{red}{$.20 \spm .01$} & \textcolor{red}{$.38 \spm .01$} & \textcolor{red}{$.00 \spm .00$} & \textcolor{red}{$1.411 \spm .025$} & \textcolor{red}{$.46 \spm .01$} & \textcolor{red}{$.093 \spm .003$} & \textcolor{red}{$.168 \spm .001$} & \textcolor{red}{$.084 \spm .001$} & \textcolor{ForestGreen}{$.043 \spm .001$} & \textcolor{ForestGreen}{$.008 \spm .001$} \\
TS (ours) & 0.1 & APS & Ada & \textcolor{red}{$.15 \spm .00$} & \textcolor{red}{$.44 \spm .25$} & $.00 \spm .00$ & \textcolor{red}{$1.349 \spm .017$} & \textcolor{red}{$.42 \spm .00$} & \textcolor{ForestGreen}{$.013 \spm .000$} & \textcolor{red}{$.003 \spm .002$} & \textcolor{red}{$.070 \spm .002$} & $.081 \spm .001$ & \textcolor{red}{$.068 \spm .001$} \\
TS (ours) & 0.1 & MSP & Const & \textcolor{ForestGreen}{$.06 \spm .00$} & \textcolor{ForestGreen}{$.14 \spm .03$} & \textcolor{ForestGreen}{$\underline{.00 \spm .00}$} & \textcolor{ForestGreen}{$1.065 \spm .009$} & \textcolor{ForestGreen}{$.39 \spm .00$} & \textcolor{ForestGreen}{$.005 \spm .000$} & \textcolor{red}{$.069 \spm .005$} & \textcolor{ForestGreen}{$.003 \spm .004$} & \textcolor{ForestGreen}{$.009 \spm .003$} & \textcolor{ForestGreen}{$\mathbf{.000 \spm .001}$} \\
TS (ours) & 0.1 & MSP & Ada & \textcolor{ForestGreen}{$.11 \spm .00$} & \textcolor{red}{$.35 \spm .02$} & \textcolor{ForestGreen}{$.00 \spm .00$} & \textcolor{ForestGreen}{$1.285 \spm .024$} & \textcolor{red}{$.42 \spm .00$} & \textcolor{ForestGreen}{$.009 \spm .000$} & \textcolor{red}{$.002 \spm .004$} & \textcolor{red}{$.071 \spm .004$} & \textcolor{ForestGreen}{$.068 \spm .003$} & \textcolor{ForestGreen}{$.054 \spm .003$} \\
TS (ours) & 0.2 & APS & Const & \textcolor{red}{$.21 \spm .01$} & \textcolor{red}{$.37 \spm .01$} & \textcolor{red}{$.00 \spm .00$} & \textcolor{red}{$1.402 \spm .017$} & \textcolor{red}{$.46 \spm .00$} & \textcolor{red}{$.092 \spm .003$} & \textcolor{red}{$.169 \spm .001$} & \textcolor{red}{$.088 \spm .000$} & \textcolor{ForestGreen}{$.043 \spm .000$} & \textcolor{ForestGreen}{$.007 \spm .000$} \\
TS (ours) & 0.2 & APS & Ada & \textcolor{ForestGreen}{$.13 \spm .00$} & \textcolor{red}{$.37 \spm .08$} & \textcolor{ForestGreen}{$.00 \spm .00$} & \textcolor{red}{$1.320 \spm .017$} & \textcolor{ForestGreen}{$.42 \spm .00$} & \textcolor{ForestGreen}{$.010 \spm .001$} & \textcolor{ForestGreen}{$\underline{.001 \spm .003}$} & \textcolor{ForestGreen}{$.047 \spm .003$} & \textcolor{ForestGreen}{$.070 \spm .002$} & \textcolor{ForestGreen}{$.062 \spm .001$} \\
TS (ours) & 0.2 & MSP & Const & \textcolor{ForestGreen}{$.06 \spm .01$} & \textcolor{red}{$.54 \spm .10$} & \textcolor{ForestGreen}{$\underline{.00 \spm .00}$} & \textcolor{ForestGreen}{$1.194 \spm .018$} & \textcolor{ForestGreen}{$.41 \spm .00$} & \textcolor{ForestGreen}{$.011 \spm .000$} & \textcolor{ForestGreen}{$\mathbf{.001 \spm .006}$} & \textcolor{ForestGreen}{$\underline{.003 \spm .005}$} & \textcolor{ForestGreen}{$.015 \spm .004$} & \textcolor{ForestGreen}{$.017 \spm .003$} \\
TS (ours) & 0.2 & MSP & Ada & \textcolor{ForestGreen}{$.12 \spm .01$} & \textcolor{red}{$.32 \spm .02$} & \textcolor{ForestGreen}{$.00 \spm .00$} & \textcolor{ForestGreen}{$1.267 \spm .014$} & \textcolor{red}{$.42 \spm .00$} & \textcolor{ForestGreen}{$.012 \spm .000$} & \textcolor{red}{$.003 \spm .002$} & \textcolor{ForestGreen}{$.030 \spm .004$} & \textcolor{ForestGreen}{$.046 \spm .004$} & \textcolor{ForestGreen}{$.041 \spm .003$} \\
\bottomrule
\end{tabular}
}
\end{table*}
  \begin{table*}[htbp]
\small
\caption{Evaluation results on CIFAR-100 and ViT-B16.
Metrics\textsuperscript{*} are scaled by \(100\). \textcolor{red}{Orange}/\textcolor{ForestGreen}{Green} indicate worse/better performance than the uncalibrated model. \textbf{Bold} and \underline{underline} mark the best and second-best values.}
\label{tab:results_cifar100_vit}
\resizebox{\textwidth}{!}{%
\begin{tabular}{@{}l @{\hspace{2pt}}l @{\hspace{2pt}}l @{\hspace{2pt}}l @{\hspace{8pt}}c c c c c c c c c c}
\toprule
Calibrator & $\alpha$ & Score & Transf. & ECE & MCE & cw-ECE & NLL & Brier Score & CMCE & $\alpha$-CMCE (0.2) & $\alpha$-CMCE (0.1) & $\alpha$-CMCE (0.05) & $\alpha$-CMCE (0.01) \\
\midrule
Base &  &  &  & $.05 \spm .00$ & $.34 \spm .09$ & $.00 \spm .00$ & $\underline{.437 \spm .002}$ & $.17 \spm .00$ & $.003 \spm .000$ & $.132 \spm .001$ & $.049 \spm .001$ & $.013 \spm .001$ & $\underline{.001 \spm .001}$ \\
\midrule
Ada-TS &  &  &  & \textcolor{red}{$.07 \spm .00$} & \textcolor{red}{$.36 \spm .06$} & $.00 \spm .00$ & \textcolor{red}{$.889 \spm .032$} & \textcolor{red}{$.18 \spm .00$} & \textcolor{red}{$.008 \spm .001$} & \textcolor{ForestGreen}{$.128 \spm .001$} & \textcolor{ForestGreen}{$.036 \spm .001$} & \textcolor{ForestGreen}{$.007 \spm .002$} & \textcolor{red}{$.035 \spm .002$} \\
Dirichlet &  &  &  & \textcolor{red}{$.13 \spm .01$} & \textcolor{red}{$.70 \spm .15$} & \textcolor{red}{$.00 \spm .00$} & \textcolor{red}{$3.134 \spm .463$} & \textcolor{red}{$.27 \spm .01$} & \textcolor{red}{$.012 \spm .003$} & \textcolor{ForestGreen}{$\mathbf{.065 \spm .007}$} & \textcolor{ForestGreen}{$\underline{.033 \spm .008}$} & \textcolor{red}{$.082 \spm .009$} & \textcolor{red}{$.118 \spm .010$} \\
Isotonic &  &  &  & \textcolor{ForestGreen}{$.04 \spm .00$} & \textcolor{ForestGreen}{$\mathbf{.21 \spm .03}$} & \textcolor{ForestGreen}{$\underline{.00 \spm .00}$} & \textcolor{red}{$1.067 \spm .116$} & \textcolor{ForestGreen}{$.16 \spm .00$} & \textcolor{red}{$.014 \spm .002$} & \textcolor{red}{$.138 \spm .003$} & \textcolor{red}{$.049 \spm .003$} & \textcolor{ForestGreen}{$.007 \spm .004$} & \textcolor{red}{$.024 \spm .004$} \\
Platt Scaling &  &  &  & \textcolor{ForestGreen}{$.04 \spm .00$} & \textcolor{red}{$.49 \spm .40$} & \textcolor{red}{$.00 \spm .00$} & \textcolor{ForestGreen}{$\mathbf{.404 \spm .006}$} & \textcolor{ForestGreen}{$\mathbf{.16 \spm .00}$} & \textcolor{red}{$.007 \spm .001$} & \textcolor{red}{$.144 \spm .002$} & \textcolor{red}{$.064 \spm .001$} & \textcolor{red}{$.033 \spm .002$} & \textcolor{red}{$.008 \spm .001$} \\
Temp. scaling &  &  &  & \textcolor{ForestGreen}{$.04 \spm .02$} & \textcolor{ForestGreen}{$.30 \spm .06$} & \textcolor{red}{$.00 \spm .00$} & \textcolor{red}{$.446 \spm .029$} & \textcolor{ForestGreen}{$\underline{.16 \spm .00}$} & \textcolor{red}{$.027 \spm .016$} & \textcolor{red}{$.159 \spm .013$} & \textcolor{red}{$.080 \spm .013$} & \textcolor{red}{$.043 \spm .006$} & \textcolor{red}{$.010 \spm .000$} \\
V.-Abers (OvA) &  &  &  & \textcolor{red}{$.10 \spm .00$} & \textcolor{ForestGreen}{$\underline{.30 \spm .04}$} & \textcolor{red}{$.00 \spm .00$} & \textcolor{red}{$.484 \spm .004$} & \textcolor{red}{$.17 \spm .00$} & \textcolor{red}{$.044 \spm .001$} & \textcolor{red}{$.183 \spm .001$} & \textcolor{red}{$.095 \spm .001$} & \textcolor{red}{$.049 \spm .000$} & \textcolor{red}{$.010 \spm .000$} \\
\midrule
Naive CMCE &  &  &  & \textcolor{red}{$.06 \spm .00$} & \textcolor{ForestGreen}{$.32 \spm .01$} & $.00 \spm .00$ & \textcolor{red}{$.456 \spm .006$} & \textcolor{red}{$.17 \spm .00$} & \textcolor{ForestGreen}{$\mathbf{.001 \spm .001}$} & \textcolor{ForestGreen}{$.126 \spm .001$} & \textcolor{ForestGreen}{$.041 \spm .001$} & \textcolor{ForestGreen}{$.005 \spm .002$} & \textcolor{red}{$.009 \spm .003$} \\
TS (ours) & 0.01 & APS & Const & \textcolor{ForestGreen}{$.04 \spm .00$} & \textcolor{red}{$.40 \spm .04$} & \textcolor{red}{$.00 \spm .00$} & \textcolor{red}{$.455 \spm .004$} & \textcolor{ForestGreen}{$.16 \spm .00$} & \textcolor{red}{$.029 \spm .002$} & \textcolor{red}{$.161 \spm .001$} & \textcolor{red}{$.081 \spm .001$} & \textcolor{red}{$.043 \spm .001$} & \textcolor{red}{$.010 \spm .000$} \\
TS (ours) & 0.01 & APS & Ada & \textcolor{red}{$.06 \spm .00$} & \textcolor{red}{$.57 \spm .31$} & $.00 \spm .00$ & \textcolor{red}{$.468 \spm .008$} & \textcolor{red}{$.17 \spm .00$} & \textcolor{ForestGreen}{$\underline{.002 \spm .001}$} & \textcolor{ForestGreen}{$.124 \spm .002$} & \textcolor{ForestGreen}{$.041 \spm .002$} & \textcolor{ForestGreen}{$.005 \spm .002$} & \textcolor{red}{$.005 \spm .001$} \\
TS (ours) & 0.01 & MSP & Const & \textcolor{red}{$.06 \spm .00$} & \textcolor{red}{$.43 \spm .09$} & $.00 \spm .00$ & \textcolor{red}{$.455 \spm .002$} & \textcolor{red}{$.17 \spm .00$} & \textcolor{red}{$.004 \spm .001$} & \textcolor{ForestGreen}{$.124 \spm .001$} & \textcolor{ForestGreen}{$.038 \spm .002$} & \textcolor{ForestGreen}{$\underline{.002 \spm .002}$} & \textcolor{ForestGreen}{$\mathbf{.000 \spm .002}$} \\
TS (ours) & 0.01 & MSP & Ada & \textcolor{red}{$.05 \spm .00$} & \textcolor{red}{$.39 \spm .14$} & \textcolor{ForestGreen}{$\underline{.00 \spm .00}$} & \textcolor{red}{$.448 \spm .004$} & \textcolor{red}{$.17 \spm .00$} & \textcolor{red}{$.004 \spm .000$} & \textcolor{ForestGreen}{$.128 \spm .001$} & \textcolor{ForestGreen}{$.044 \spm .001$} & \textcolor{ForestGreen}{$.007 \spm .002$} & \textcolor{red}{$.004 \spm .003$} \\
TS (ours) & 0.05 & APS & Const & \textcolor{red}{$.10 \spm .00$} & \textcolor{red}{$.45 \spm .03$} & \textcolor{red}{$.00 \spm .00$} & \textcolor{red}{$.562 \spm .003$} & \textcolor{red}{$.18 \spm .00$} & \textcolor{red}{$.066 \spm .001$} & \textcolor{red}{$.184 \spm .001$} & \textcolor{red}{$.096 \spm .001$} & \textcolor{red}{$.050 \spm .000$} & \textcolor{red}{$.010 \spm .000$} \\
TS (ours) & 0.05 & APS & Ada & \textcolor{red}{$.06 \spm .01$} & \textcolor{red}{$.68 \spm .22$} & $.00 \spm .00$ & \textcolor{red}{$.486 \spm .014$} & \textcolor{red}{$.18 \spm .00$} & \textcolor{red}{$.006 \spm .006$} & \textcolor{ForestGreen}{$.126 \spm .008$} & \textcolor{ForestGreen}{$.041 \spm .006$} & \textcolor{ForestGreen}{$.003 \spm .005$} & \textcolor{red}{$.002 \spm .001$} \\
TS (ours) & 0.05 & MSP & Const & \textcolor{ForestGreen}{$.04 \spm .00$} & \textcolor{red}{$.36 \spm .19$} & \textcolor{ForestGreen}{$\underline{.00 \spm .00}$} & \textcolor{red}{$.456 \spm .002$} & \textcolor{red}{$.17 \spm .00$} & \textcolor{red}{$.014 \spm .000$} & \textcolor{ForestGreen}{$.122 \spm .000$} & \textcolor{ForestGreen}{$.040 \spm .001$} & \textcolor{ForestGreen}{$\mathbf{.001 \spm .002}$} & \textcolor{red}{$.009 \spm .000$} \\
TS (ours) & 0.05 & MSP & Ada & \textcolor{ForestGreen}{$\underline{.03 \spm .00}$} & \textcolor{red}{$.38 \spm .06$} & \textcolor{ForestGreen}{$\mathbf{.00 \spm .00}$} & \textcolor{red}{$.472 \spm .005$} & \textcolor{red}{$.17 \spm .00$} & \textcolor{red}{$.013 \spm .001$} & \textcolor{ForestGreen}{$.126 \spm .002$} & \textcolor{ForestGreen}{$.036 \spm .003$} & \textcolor{ForestGreen}{$.007 \spm .004$} & \textcolor{red}{$.005 \spm .001$} \\
TS (ours) & 0.1 & APS & Const & \textcolor{red}{$.16 \spm .00$} & \textcolor{red}{$.46 \spm .02$} & \textcolor{red}{$.00 \spm .00$} & \textcolor{red}{$.648 \spm .004$} & \textcolor{red}{$.20 \spm .00$} & \textcolor{red}{$.091 \spm .001$} & \textcolor{red}{$.189 \spm .000$} & \textcolor{red}{$.099 \spm .000$} & \textcolor{red}{$.050 \spm .000$} & \textcolor{red}{$.010 \spm .000$} \\
TS (ours) & 0.1 & APS & Ada & \textcolor{red}{$.12 \spm .09$} & \textcolor{red}{$.60 \spm .16$} & \textcolor{red}{$.00 \spm .00$} & \textcolor{red}{$.619 \spm .136$} & \textcolor{red}{$.21 \spm .04$} & \textcolor{red}{$.058 \spm .045$} & \textcolor{red}{$.149 \spm .014$} & \textcolor{red}{$.060 \spm .013$} & \textcolor{red}{$.028 \spm .008$} & \textcolor{red}{$.002 \spm .004$} \\
TS (ours) & 0.1 & MSP & Const & \textcolor{ForestGreen}{$\mathbf{.01 \spm .00}$} & \textcolor{red}{$.40 \spm .12$} & \textcolor{ForestGreen}{$\underline{.00 \spm .00}$} & \textcolor{red}{$.543 \spm .007$} & \textcolor{red}{$.18 \spm .00$} & \textcolor{red}{$.029 \spm .000$} & \textcolor{ForestGreen}{$\underline{.109 \spm .001}$} & \textcolor{ForestGreen}{$\mathbf{.009 \spm .001}$} & \textcolor{red}{$.032 \spm .001$} & \textcolor{red}{$.002 \spm .001$} \\
TS (ours) & 0.1 & MSP & Ada & \textcolor{red}{$.07 \spm .00$} & \textcolor{red}{$.34 \spm .08$} & \textcolor{red}{$.00 \spm .00$} & \textcolor{red}{$.484 \spm .004$} & \textcolor{red}{$.17 \spm .00$} & \textcolor{red}{$.030 \spm .000$} & \textcolor{ForestGreen}{$.129 \spm .000$} & \textcolor{ForestGreen}{$.043 \spm .001$} & \textcolor{red}{$.036 \spm .002$} & \textcolor{red}{$.009 \spm .000$} \\
TS (ours) & 0.2 & APS & Const & \textcolor{red}{$.22 \spm .00$} & \textcolor{red}{$.60 \spm .03$} & \textcolor{red}{$.01 \spm .00$} & \textcolor{red}{$.760 \spm .002$} & \textcolor{red}{$.25 \spm .00$} & \textcolor{red}{$.116 \spm .001$} & \textcolor{red}{$.185 \spm .001$} & \textcolor{red}{$.094 \spm .000$} & \textcolor{red}{$.048 \spm .000$} & \textcolor{red}{$.009 \spm .000$} \\
TS (ours) & 0.2 & APS & Ada & \textcolor{red}{$.13 \spm .06$} & \textcolor{red}{$.62 \spm .15$} & \textcolor{red}{$.00 \spm .00$} & \textcolor{red}{$.629 \spm .090$} & \textcolor{red}{$.21 \spm .03$} & \textcolor{red}{$.066 \spm .030$} & \textcolor{red}{$.155 \spm .008$} & \textcolor{red}{$.076 \spm .005$} & \textcolor{red}{$.032 \spm .005$} & \textcolor{red}{$.003 \spm .002$} \\
TS (ours) & 0.2 & MSP & Const & \textcolor{red}{$.16 \spm .00$} & $.34 \spm .09$ & \textcolor{red}{$.00 \spm .00$} & \textcolor{red}{$.556 \spm .002$} & \textcolor{red}{$.19 \spm .00$} & \textcolor{red}{$.064 \spm .000$} & $.132 \spm .001$ & \textcolor{red}{$.078 \spm .001$} & \textcolor{red}{$.042 \spm .001$} & \textcolor{red}{$.010 \spm .000$} \\
TS (ours) & 0.2 & MSP & Ada & \textcolor{red}{$.14 \spm .01$} & \textcolor{red}{$.37 \spm .08$} & \textcolor{red}{$.00 \spm .00$} & \textcolor{red}{$.554 \spm .007$} & \textcolor{red}{$.19 \spm .00$} & \textcolor{red}{$.060 \spm .002$} & \textcolor{ForestGreen}{$.125 \spm .002$} & \textcolor{red}{$.076 \spm .004$} & \textcolor{red}{$.037 \spm .002$} & \textcolor{red}{$.008 \spm .001$} \\
\bottomrule
\end{tabular}
}
\end{table*}

  \begin{table*}[htbp]
\caption{Evaluation results on CIFAR-100 and ViT-B16 model where our calibrators have $\alpha = 0.01$, MSP nonconformity score and constant transformation. Metrics\textsuperscript{*} are scaled by \(100\). \textcolor{red}{Orange}/\textcolor{ForestGreen}{Green} indicate worse/better performance than the uncalibrated model. \textbf{Bold} and \underline{underline} mark the best and second-best values.}
\label{tab:results_cifar100_vit_best}
\resizebox{\textwidth}{!}{%
\begin{tabular}{@{}l @{\hspace{8pt}}c c c c c c c}
\toprule
Calibrator & ECE & MCE & cw-ECE & NLL & Brier Score & CMCE & $\alpha$-CMCE (0.01) \\
\midrule
Base
& $.05 \spm .00$ & $.34 \spm .09$ & $\underline{.00 \spm .00}$ & $\underline{.437 \spm .002}$ & $.17 \spm .00$ & $\underline{.003 \spm .000}$ & $\underline{.001 \spm .001}$ \\
\midrule
Ada-TS
& \textcolor{red}{$.07 \spm .00$} & \textcolor{red}{$.36 \spm .06$} & $\underline{.00 \spm .00}$ & \textcolor{red}{$.889 \spm .032$} & \textcolor{red}{$.18 \spm .00$} & \textcolor{red}{$.008 \spm .001$} & \textcolor{red}{$.035 \spm .002$} \\
Dirichlet
& \textcolor{red}{$.13 \spm .01$} & \textcolor{red}{$.70 \spm .15$} & \textcolor{red}{$.00 \spm .00$} & \textcolor{red}{$3.134 \spm .463$} & \textcolor{red}{$.27 \spm .01$} & \textcolor{red}{$.012 \spm .003$} & \textcolor{red}{$.118 \spm .010$} \\
Isotonic
& \textcolor{ForestGreen}{$\underline{.04 \spm .00}$} & \textcolor{ForestGreen}{$\mathbf{.21 \spm .03}$} & \textcolor{ForestGreen}{$\mathbf{.00 \spm .00}$} & \textcolor{red}{$1.067 \spm .116$} & \textcolor{ForestGreen}{$.16 \spm .00$} & \textcolor{red}{$.014 \spm .002$} & \textcolor{red}{$.024 \spm .004$} \\
Platt Scaling
& \textcolor{ForestGreen}{$\mathbf{.04 \spm .00}$} & \textcolor{red}{$.49 \spm .40$} & \textcolor{red}{$.00 \spm .00$} & \textcolor{ForestGreen}{$\mathbf{.404 \spm .006}$} & \textcolor{ForestGreen}{$\mathbf{.16 \spm .00}$} & \textcolor{red}{$.007 \spm .001$} & \textcolor{red}{$.008 \spm .001$} \\
Temp. scaling
& \textcolor{ForestGreen}{$.04 \spm .02$} & \textcolor{ForestGreen}{$.30 \spm .06$} & \textcolor{red}{$.00 \spm .00$} & \textcolor{red}{$.446 \spm .029$} & \textcolor{ForestGreen}{$\underline{.16 \spm .00}$} & \textcolor{red}{$.027 \spm .016$} & \textcolor{red}{$.010 \spm .000$} \\
V.-Abers (OvA)
& \textcolor{red}{$.10 \spm .00$} & \textcolor{ForestGreen}{$\underline{.30 \spm .04}$} & \textcolor{red}{$.00 \spm .00$} & \textcolor{red}{$.484 \spm .004$} & \textcolor{red}{$.17 \spm .00$} & \textcolor{red}{$.044 \spm .001$} & \textcolor{red}{$.010 \spm .000$} \\
\midrule
Naive CMCE
& \textcolor{red}{$.06 \spm .00$} & \textcolor{ForestGreen}{$.32 \spm .01$} & $\underline{.00 \spm .00}$ & \textcolor{red}{$.456 \spm .006$} & \textcolor{red}{$.17 \spm .00$} & \textcolor{ForestGreen}{$\mathbf{.001 \spm .001}$} & \textcolor{red}{$.009 \spm .003$} \\
TS (ours)
& \textcolor{red}{$.06 \spm .00$} & \textcolor{red}{$.43 \spm .09$} & $\underline{.00 \spm .00}$ & \textcolor{red}{$.455 \spm .002$} & \textcolor{red}{$.17 \spm .00$} & \textcolor{red}{$.004 \spm .001$} & \textcolor{ForestGreen}{$\mathbf{.000 \spm .002}$} \\
\bottomrule
\end{tabular}
}
\end{table*}
  




\subsection{ImageNet}
\begin{table*}[htbp]
\centering
\caption{Evaluation results on ImageNet with calibrators configured using $\alpha = 0.01$, the APS nonconformity score, and adaptive transformation. Metrics\textsuperscript{*} denote values scaled by a factor of $100$. \textcolor{red}{Red} indicates performance worse than the uncalibrated model, whereas \textcolor{ForestGreen}{Green} indicates performance better than the uncalibrated model. \textbf{Bold} highlights the best value, and \underline{underline} highlights the second-best value.
}
\label{tab:results_imagenet_best}
\resizebox{\textwidth}{!}{%
\begin{tabular}{@{}l @{\hspace{8pt}}c c c c c c}
\toprule
Calibrator & ECE\textsuperscript{*} & MCE\textsuperscript{*} & NLL & Brier Score\textsuperscript{*} & CMCE\textsuperscript{*} & $\alpha$-CMCE (0.01)\textsuperscript{*} \\
\midrule
Base
& $\mathbf{2.41 \spm 0.05}$ & $\mathbf{8.92 \spm 1.23}$ & $\mathbf{0.70 \spm 0.00}$ & $\mathbf{26.06 \spm 0.05}$ & $0.43 \spm 0.01$ & $0.40 \spm 0.02$ \\
\midrule
Ada-TS
& \textcolor{red}{$10.81 \spm 0.41$} & \textcolor{red}{$25.53 \spm 1.31$} & \textcolor{red}{$1.69 \spm 0.10$} & \textcolor{red}{$30.45 \spm 0.30$} & \textcolor{red}{$0.58 \spm 0.10$} & \textcolor{red}{$7.21 \spm 0.48$} \\
Isotonic
& \textcolor{red}{$4.43 \spm 0.08$} & \textcolor{red}{$11.35 \spm 1.00$} & \textcolor{red}{$2.12 \spm 0.02$} & \textcolor{red}{$27.36 \spm 0.15$} & \textcolor{red}{$2.56 \spm 0.09$} & \textcolor{red}{$5.71 \spm 0.09$} \\
Platt Scaling
& \textcolor{red}{$12.89 \spm 0.19$} & \textcolor{red}{$20.67 \spm 1.23$} & \textcolor{red}{$0.81 \spm 0.00$} & \textcolor{red}{$28.62 \spm 0.07$} & \textcolor{red}{$1.26 \spm 0.02$} & \textcolor{red}{$0.92 \spm 0.02$} \\
Temp. scaling
& \textcolor{red}{$\underline{2.49 \spm 0.04}$} & \textcolor{red}{$\underline{9.08 \spm 1.04}$} & $\mathbf{0.70 \spm 0.00}$ & $\mathbf{26.06 \spm 0.05}$ & \textcolor{ForestGreen}{$0.42 \spm 0.01$} & \textcolor{ForestGreen}{$\underline{0.39 \spm 0.03}$} \\
V.-Abers (OvA)
& \textcolor{red}{$23.93 \spm 0.25$} & \textcolor{red}{$33.66 \spm 0.77$} & \textcolor{red}{$1.00 \spm 0.00$} & \textcolor{red}{$33.32 \spm 0.14$} & \textcolor{red}{$7.92 \spm 0.04$} & \textcolor{red}{$1.00 \spm 0.00$} \\
\midrule
Naive CMCE
& \textcolor{red}{$5.82 \spm 0.09$} & \textcolor{red}{$16.18 \spm 0.99$} & \textcolor{red}{$0.73 \spm 0.00$} & \textcolor{red}{$26.72 \spm 0.07$} & \textcolor{ForestGreen}{$\mathbf{0.02 \spm 0.00}$} & \textcolor{red}{$0.93 \spm 0.06$} \\
TS (ours)
& \textcolor{red}{$2.87 \spm 0.06$} & \textcolor{red}{$13.34 \spm 9.96$} & \textcolor{red}{$\underline{0.73 \spm 0.00}$} & \textcolor{red}{$\underline{26.35 \spm 0.07}$} & \textcolor{ForestGreen}{$\underline{0.34 \spm 0.01}$} & \textcolor{ForestGreen}{$\mathbf{0.01 \spm 0.04}$} \\
\bottomrule
\end{tabular}
}
\end{table*}
  In Table~\ref{tab:results_imagenet}, we present the full results for ImageNet (ViT), and in Table~\ref{tab:results_imagenet_best}, we display the best ones.
  From the results, we see that most of the methods do not the uncalibrated baseline across all metrics except cumulative mass ones. Our methods show better performance compared with the baselines calibrations.

  Table~\ref{tab:results_imagenet} indicates that standard baselines frequently degrade calibration on this many-class task (e.g., Venn-Abers and Platt increase ECE/MCE substantially). The uncalibrated model achieves a strong NLL/Brier score.
  
  Still, our methods substantially reduce CMCE: e.g., TS (MSP, Const, \(\alpha=0.01\)) achieves the second-best CMCE among calibrated models and improves ECE/MCE relative to most competitors, while maintaining competitive coverage near both targets. Moreover, all of  our methods improve $\alpha$-CMCE for $\alpha$'s used during the calibration procedure.

      

  \begin{table*}[htbp]
\small
\caption{Evaluation results on Imagenet. textcolor{red}{Orange}/\textcolor{ForestGreen}{Green} indicate worse/better performance than the uncalibrated model. \textbf{Bold} and \underline{underline} mark the best and second-best values.}
\label{tab:results_imagenet}
\resizebox{\textwidth}{!}{%
\begin{tabular}{@{}l @{\hspace{2pt}}l @{\hspace{2pt}}l @{\hspace{2pt}}l @{\hspace{8pt}}c c c c c c c c c c}
\toprule
Calibrator & $\alpha$ & Score & Transf. & ECE & MCE & cw-ECE & NLL & Brier Score & CMCE & $\alpha$-CMCE (0.2) & $\alpha$-CMCE (0.1) & $\alpha$-CMCE (0.05) & $\alpha$-CMCE (0.01) \\
\midrule
Base &  &  &  & $\underline{.02 \spm .00}$ & $.09 \spm .01$ & $\mathbf{.00 \spm .00}$ & $\mathbf{.704 \spm .002}$ & $\mathbf{.26 \spm .00}$ & $.004 \spm .000$ & $.131 \spm .001$ & $.059 \spm .001$ & $.026 \spm .001$ & $.004 \spm .000$ \\
\midrule
Ada-TS &  &  &  & \textcolor{red}{$.11 \spm .00$} & \textcolor{red}{$.26 \spm .01$} & $\mathbf{.00 \spm .00}$ & \textcolor{red}{$1.688 \spm .096$} & \textcolor{red}{$.30 \spm .00$} & \textcolor{red}{$.006 \spm .001$} & \textcolor{ForestGreen}{$\mathbf{.069 \spm .004}$} & \textcolor{ForestGreen}{$.014 \spm .004$} & \textcolor{red}{$.052 \spm .004$} & \textcolor{red}{$.072 \spm .005$} \\
Isotonic &  &  &  & \textcolor{red}{$.04 \spm .00$} & \textcolor{red}{$.11 \spm .01$} & $\mathbf{.00 \spm .00}$ & \textcolor{red}{$2.124 \spm .025$} & \textcolor{red}{$.27 \spm .00$} & \textcolor{red}{$.026 \spm .001$} & \textcolor{ForestGreen}{$.089 \spm .002$} & \textcolor{ForestGreen}{$\underline{.010 \spm .001}$} & \textcolor{red}{$.029 \spm .001$} & \textcolor{red}{$.057 \spm .001$} \\
Platt Scaling &  &  &  & \textcolor{red}{$.13 \spm .00$} & \textcolor{red}{$.21 \spm .01$} & \textcolor{red}{$.00 \spm .00$} & \textcolor{red}{$.806 \spm .004$} & \textcolor{red}{$.29 \spm .00$} & \textcolor{red}{$.013 \spm .000$} & \textcolor{red}{$.167 \spm .001$} & \textcolor{red}{$.087 \spm .000$} & \textcolor{red}{$.047 \spm .000$} & \textcolor{red}{$.009 \spm .000$} \\
Temp. scaling &  &  &  & \textcolor{red}{$.02 \spm .00$} & \textcolor{red}{$.09 \spm .01$} & $\mathbf{.00 \spm .00}$ & $\mathbf{.704 \spm .002}$ & $\mathbf{.26 \spm .00}$ & \textcolor{ForestGreen}{$.004 \spm .000$} & \textcolor{ForestGreen}{$.130 \spm .001$} & \textcolor{ForestGreen}{$.058 \spm .001$} & \textcolor{ForestGreen}{$.025 \spm .001$} & \textcolor{ForestGreen}{$.004 \spm .000$} \\
V.-Abers (OvA) &  &  &  & \textcolor{red}{$.24 \spm .00$} & \textcolor{red}{$.34 \spm .01$} & \textcolor{red}{$.00 \spm .00$} & \textcolor{red}{$1.001 \spm .005$} & \textcolor{red}{$.33 \spm .00$} & \textcolor{red}{$.079 \spm .000$} & \textcolor{red}{$.195 \spm .000$} & \textcolor{red}{$.100 \spm .000$} & \textcolor{red}{$.050 \spm .000$} & \textcolor{red}{$.010 \spm .000$} \\
\midrule
Naive CMCE &  &  &  & \textcolor{red}{$.06 \spm .00$} & \textcolor{red}{$.16 \spm .01$} & $\mathbf{.00 \spm .00}$ & \textcolor{red}{$.734 \spm .003$} & \textcolor{red}{$.27 \spm .00$} & \textcolor{ForestGreen}{$\mathbf{.000 \spm .000}$} & \textcolor{ForestGreen}{$.102 \spm .001$} & \textcolor{ForestGreen}{$.031 \spm .001$} & \textcolor{ForestGreen}{$\underline{.002 \spm .000}$} & \textcolor{red}{$.009 \spm .001$} \\
TS (ours) & 0.01 & APS & Const & \textcolor{red}{$.16 \spm .00$} & \textcolor{red}{$.41 \spm .01$} & \textcolor{red}{$\underline{.00 \spm .00}$} & \textcolor{red}{$1.104 \spm .009$} & \textcolor{red}{$.32 \spm .00$} & \textcolor{red}{$.073 \spm .001$} & \textcolor{red}{$.186 \spm .001$} & \textcolor{red}{$.095 \spm .000$} & \textcolor{red}{$.048 \spm .000$} & \textcolor{red}{$.010 \spm .000$} \\
TS (ours) & 0.01 & APS & Ada & \textcolor{red}{$.03 \spm .00$} & \textcolor{red}{$.13 \spm .10$} & $\mathbf{.00 \spm .00}$ & \textcolor{red}{$.727 \spm .003$} & \textcolor{red}{$.26 \spm .00$} & \textcolor{ForestGreen}{$.003 \spm .000$} & \textcolor{ForestGreen}{$.123 \spm .001$} & \textcolor{ForestGreen}{$.052 \spm .001$} & \textcolor{ForestGreen}{$.020 \spm .001$} & \textcolor{ForestGreen}{$\mathbf{.000 \spm .000}$} \\
TS (ours) & 0.01 & MSP & Const & \textcolor{red}{$.04 \spm .00$} & \textcolor{red}{$.15 \spm .00$} & $\mathbf{.00 \spm .00}$ & \textcolor{red}{$.723 \spm .003$} & \textcolor{red}{$.27 \spm .00$} & \textcolor{ForestGreen}{$\underline{.001 \spm .000}$} & \textcolor{ForestGreen}{$.099 \spm .001$} & \textcolor{ForestGreen}{$.029 \spm .001$} & \textcolor{ForestGreen}{$.004 \spm .001$} & \textcolor{ForestGreen}{$.001 \spm .001$} \\
TS (ours) & 0.01 & MSP & Ada & \textcolor{red}{$.03 \spm .00$} & \textcolor{ForestGreen}{$\underline{.09 \spm .01}$} & $\mathbf{.00 \spm .00}$ & \textcolor{red}{$\underline{.720 \spm .005}$} & \textcolor{red}{$\underline{.26 \spm .00}$} & \textcolor{red}{$.004 \spm .000$} & \textcolor{ForestGreen}{$.129 \spm .000$} & \textcolor{ForestGreen}{$.056 \spm .001$} & \textcolor{ForestGreen}{$.022 \spm .001$} & \textcolor{ForestGreen}{$.002 \spm .001$} \\
TS (ours) & 0.05 & APS & Const & \textcolor{red}{$.22 \spm .00$} & \textcolor{red}{$.49 \spm .01$} & \textcolor{red}{$.00 \spm .00$} & \textcolor{red}{$1.244 \spm .007$} & \textcolor{red}{$.35 \spm .00$} & \textcolor{red}{$.093 \spm .001$} & \textcolor{red}{$.188 \spm .000$} & \textcolor{red}{$.097 \spm .000$} & \textcolor{red}{$.049 \spm .000$} & \textcolor{red}{$.010 \spm .000$} \\
TS (ours) & 0.05 & APS & Ada & \textcolor{red}{$.04 \spm .00$} & \textcolor{red}{$.28 \spm .15$} & $\mathbf{.00 \spm .00}$ & \textcolor{red}{$.756 \spm .007$} & \textcolor{red}{$.27 \spm .00$} & \textcolor{ForestGreen}{$.003 \spm .000$} & \textcolor{ForestGreen}{$.114 \spm .002$} & \textcolor{ForestGreen}{$.042 \spm .003$} & \textcolor{ForestGreen}{$.010 \spm .003$} & \textcolor{ForestGreen}{$.002 \spm .001$} \\
TS (ours) & 0.05 & MSP & Const & \textcolor{red}{$.04 \spm .00$} & \textcolor{red}{$.17 \spm .03$} & $\mathbf{.00 \spm .00}$ & \textcolor{red}{$.754 \spm .004$} & \textcolor{red}{$.27 \spm .00$} & \textcolor{red}{$.006 \spm .000$} & \textcolor{ForestGreen}{$.101 \spm .001$} & \textcolor{ForestGreen}{$.031 \spm .002$} & \textcolor{ForestGreen}{$\mathbf{.002 \spm .003}$} & \textcolor{ForestGreen}{$.000 \spm .001$} \\
TS (ours) & 0.05 & MSP & Ada & \textcolor{ForestGreen}{$\mathbf{.02 \spm .00}$} & \textcolor{ForestGreen}{$\mathbf{.09 \spm .01}$} & $\mathbf{.00 \spm .00}$ & \textcolor{red}{$.727 \spm .005$} & \textcolor{red}{$.26 \spm .00$} & \textcolor{red}{$.008 \spm .000$} & \textcolor{ForestGreen}{$.127 \spm .002$} & \textcolor{ForestGreen}{$.052 \spm .002$} & \textcolor{ForestGreen}{$.017 \spm .003$} & \textcolor{red}{$.004 \spm .001$} \\
TS (ours) & 0.1 & APS & Const & \textcolor{red}{$.24 \spm .00$} & \textcolor{red}{$.52 \spm .00$} & \textcolor{red}{$.00 \spm .00$} & \textcolor{red}{$1.267 \spm .007$} & \textcolor{red}{$.36 \spm .00$} & \textcolor{red}{$.097 \spm .001$} & \textcolor{red}{$.187 \spm .000$} & \textcolor{red}{$.096 \spm .000$} & \textcolor{red}{$.049 \spm .000$} & \textcolor{red}{$.010 \spm .000$} \\
TS (ours) & 0.1 & APS & Ada & \textcolor{red}{$.08 \spm .09$} & \textcolor{red}{$.35 \spm .12$} & \textcolor{red}{$\underline{.00 \spm .00}$} & \textcolor{red}{$.869 \spm .216$} & \textcolor{red}{$.29 \spm .04$} & \textcolor{red}{$.022 \spm .041$} & \textcolor{ForestGreen}{$.126 \spm .034$} & \textcolor{ForestGreen}{$.048 \spm .026$} & \textcolor{ForestGreen}{$.023 \spm .014$} & \textcolor{ForestGreen}{$\underline{.000 \spm .005}$} \\
TS (ours) & 0.1 & MSP & Const & \textcolor{red}{$.04 \spm .00$} & \textcolor{red}{$.34 \spm .03$} & $\mathbf{.00 \spm .00}$ & \textcolor{red}{$.841 \spm .007$} & \textcolor{red}{$.28 \spm .00$} & \textcolor{red}{$.013 \spm .000$} & \textcolor{ForestGreen}{$.097 \spm .001$} & \textcolor{ForestGreen}{$\mathbf{.003 \spm .002}$} & \textcolor{ForestGreen}{$.003 \spm .001$} & \textcolor{red}{$.010 \spm .001$} \\
TS (ours) & 0.1 & MSP & Ada & \textcolor{red}{$.04 \spm .00$} & \textcolor{red}{$.09 \spm .01$} & $\mathbf{.00 \spm .00}$ & \textcolor{red}{$.743 \spm .002$} & \textcolor{red}{$.27 \spm .00$} & \textcolor{red}{$.014 \spm .000$} & \textcolor{ForestGreen}{$.126 \spm .001$} & \textcolor{ForestGreen}{$.051 \spm .001$} & \textcolor{red}{$.030 \spm .000$} & \textcolor{red}{$.006 \spm .000$} \\
TS (ours) & 0.2 & APS & Const & \textcolor{red}{$.24 \spm .00$} & \textcolor{red}{$.52 \spm .00$} & \textcolor{red}{$.00 \spm .00$} & \textcolor{red}{$1.263 \spm .005$} & \textcolor{red}{$.37 \spm .00$} & \textcolor{red}{$.098 \spm .001$} & \textcolor{red}{$.185 \spm .000$} & \textcolor{red}{$.097 \spm .000$} & \textcolor{red}{$.049 \spm .000$} & \textcolor{red}{$.009 \spm .000$} \\
TS (ours) & 0.2 & APS & Ada & \textcolor{red}{$.03 \spm .00$} & \textcolor{red}{$.38 \spm .05$} & $\mathbf{.00 \spm .00}$ & \textcolor{red}{$.785 \spm .003$} & \textcolor{red}{$.27 \spm .00$} & \textcolor{red}{$.007 \spm .000$} & \textcolor{ForestGreen}{$.108 \spm .002$} & \textcolor{ForestGreen}{$.054 \spm .002$} & \textcolor{ForestGreen}{$.019 \spm .001$} & \textcolor{ForestGreen}{$.002 \spm .000$} \\
TS (ours) & 0.2 & MSP & Const & \textcolor{red}{$.07 \spm .00$} & \textcolor{red}{$.12 \spm .02$} & $\mathbf{.00 \spm .00}$ & \textcolor{red}{$.878 \spm .003$} & \textcolor{red}{$.29 \spm .00$} & \textcolor{red}{$.033 \spm .000$} & \textcolor{ForestGreen}{$\underline{.075 \spm .001}$} & \textcolor{red}{$.061 \spm .001$} & \textcolor{ForestGreen}{$.025 \spm .001$} & \textcolor{ForestGreen}{$.002 \spm .001$} \\
TS (ours) & 0.2 & MSP & Ada & \textcolor{red}{$.09 \spm .00$} & \textcolor{red}{$.14 \spm .00$} & \textcolor{red}{$\underline{.00 \spm .00}$} & \textcolor{red}{$.805 \spm .003$} & \textcolor{red}{$.28 \spm .00$} & \textcolor{red}{$.030 \spm .000$} & \textcolor{ForestGreen}{$.120 \spm .001$} & \textcolor{red}{$.067 \spm .001$} & \textcolor{red}{$.032 \spm .001$} & \textcolor{red}{$.005 \spm .001$} \\
\bottomrule
\end{tabular}
}
\end{table*}  

\subsection{iNaturalist21}

  Finally, in Table~\ref{tab:results_inaturalist} we present the best results for iNaturalist21, obtained with ViT, and in Table~\ref{tab:results_cifar100_inaturalist_best} we present the best results for our model.

  Table~\ref{tab:results_inaturalist} shows that OvR-style binary calibration methods (Isotonic, Platt) harm accuracy and worsen several key calibration metrics on this dataset with a large number of classes. 
  By contrast, our methods preserve accuracy and improve NLL, Brier, and especially CMCE, while meeting coverage targets more reliably. Moreover, a particular instance is the best across all key calibration metrics (see Table~\ref{tab:results_cifar100_inaturalist_best}).
  Temperature Scaling is competitive on NLL/Brier but is worse than our methods in terms of CMCE and target coverage.  
  
  \begin{table*}[htbp]
\centering
\caption{Calibration results on iNaturalist21 (10k classes). Binary OvR methods often reduce accuracy and worsen calibration, whereas our proposed approaches preserve accuracy and achieve superior CMCE, coverage, and other key calibration metrics.}
\label{tab:results_inaturalist}
\resizebox{\textwidth}{!}{%
\begin{tabular}{@{}l @{\hspace{2pt}}l @{\hspace{2pt}}l @{\hspace{2pt}}l @{\hspace{8pt}}c c c c c c c c c c c}
\toprule
Calibrator & $\alpha$ & Score & Transf. & Accuracy & ECE & MCE & cw-ECE & NLL & Brier Score & CMCE & $\alpha$-CMCE (0.2) & $\alpha$-CMCE (0.1) & $\alpha$-CMCE (0.05) & $\alpha$-CMCE (0.01) \\
\midrule
Base &  &  &  & $\mathbf{.78 \spm .00}$ & $.50 \spm .00$ & $.65 \spm .01$ & $\underline{.0001 \spm .0000}$ & $2.043 \spm .015$ & $.61 \spm .00$ & $.056 \spm .000$ & $.188 \spm .001$ & $.096 \spm .000$ & $.048 \spm .000$ & $.010 \spm .000$ \\
\midrule
Ada-TS &  &  &  & $\mathbf{.78 \spm .00}$ & \textcolor{ForestGreen}{$.11 \spm .01$} & \textcolor{ForestGreen}{$.27 \spm .02$} & \textcolor{ForestGreen}{$\mathbf{.0000 \spm .0000}$} & \textcolor{red}{$2.238 \spm .098$} & \textcolor{ForestGreen}{$.36 \spm .00$} & \textcolor{ForestGreen}{$.009 \spm .001$} & \textcolor{ForestGreen}{$\mathbf{.052 \spm .007}$} & \textcolor{ForestGreen}{$.030 \spm .007$} & \textcolor{red}{$.067 \spm .007$} & \textcolor{red}{$.086 \spm .006$} \\
Isotonic &  &  &  & \textcolor{red}{$.45 \spm .00$} & \textcolor{ForestGreen}{$.26 \spm .01$} & \textcolor{ForestGreen}{$.38 \spm .01$} & $\underline{.0001 \spm .0000}$ & \textcolor{red}{$13.707 \spm .119$} & \textcolor{red}{$.80 \spm .01$} & \textcolor{red}{$.175 \spm .006$} & \textcolor{red}{$.309 \spm .004$} & \textcolor{red}{$.400 \spm .005$} & \textcolor{red}{$.445 \spm .005$} & \textcolor{red}{$.480 \spm .004$} \\
Platt Scaling &  &  &  & \textcolor{red}{$\underline{.47 \spm .00}$} & \textcolor{ForestGreen}{$.14 \spm .00$} & \textcolor{ForestGreen}{$.30 \spm .00$} & $\underline{.0001 \spm .0000}$ & \textcolor{red}{$12.506 \spm .090$} & \textcolor{red}{$.68 \spm .00$} & \textcolor{red}{$.246 \spm .003$} & \textcolor{red}{$.232 \spm .003$} & \textcolor{red}{$.330 \spm .003$} & \textcolor{red}{$.379 \spm .003$} & \textcolor{red}{$.418 \spm .003$} \\
Temp. scaling &  &  &  & $\mathbf{.78 \spm .00}$ & \textcolor{ForestGreen}{$.13 \spm .08$} & \textcolor{ForestGreen}{$.26 \spm .11$} & $\underline{.0001 \spm .0000}$ & \textcolor{ForestGreen}{$1.293 \spm .100$} & \textcolor{ForestGreen}{$.36 \spm .04$} & \textcolor{ForestGreen}{$.003 \spm .004$} & \textcolor{ForestGreen}{$.121 \spm .021$} & \textcolor{ForestGreen}{$.046 \spm .019$} & \textcolor{ForestGreen}{$.013 \spm .016$} & \textcolor{ForestGreen}{$.004 \spm .009$} \\
\midrule
Naive CMCE &  &  &  & $\mathbf{.78 \spm .00}$ & \textcolor{ForestGreen}{$.13 \spm .04$} & \textcolor{ForestGreen}{$.26 \spm .05$} & $\underline{.0001 \spm .0000}$ & \textcolor{ForestGreen}{$1.274 \spm .026$} & \textcolor{ForestGreen}{$.35 \spm .01$} & \textcolor{ForestGreen}{$.001 \spm .000$} & \textcolor{ForestGreen}{$.121 \spm .011$} & \textcolor{ForestGreen}{$.048 \spm .009$} & \textcolor{ForestGreen}{$.014 \spm .009$} & \textcolor{ForestGreen}{$.003 \spm .006$} \\
TS (ours) & 0.01 & APS & Const & $\mathbf{.78 \spm .00}$ & \textcolor{ForestGreen}{$.26 \spm .01$} & \textcolor{ForestGreen}{$.39 \spm .02$} & $\underline{.0001 \spm .0000}$ & \textcolor{ForestGreen}{$1.427 \spm .020$} & \textcolor{ForestGreen}{$.41 \spm .01$} & \textcolor{ForestGreen}{$.007 \spm .001$} & \textcolor{ForestGreen}{$.148 \spm .003$} & \textcolor{ForestGreen}{$.071 \spm .002$} & \textcolor{ForestGreen}{$.035 \spm .002$} & \textcolor{ForestGreen}{$.007 \spm .001$} \\
TS (ours) & 0.01 & APS & Ada & $\mathbf{.78 \spm .00}$ & \textcolor{ForestGreen}{$.44 \spm .02$} & \textcolor{ForestGreen}{$.61 \spm .01$} & $\underline{.0001 \spm .0000}$ & \textcolor{ForestGreen}{$1.920 \spm .050$} & \textcolor{ForestGreen}{$.57 \spm .01$} & \textcolor{ForestGreen}{$.047 \spm .002$} & \textcolor{ForestGreen}{$.177 \spm .003$} & \textcolor{ForestGreen}{$.087 \spm .002$} & \textcolor{ForestGreen}{$.042 \spm .001$} & \textcolor{ForestGreen}{$.006 \spm .001$} \\
TS (ours) & 0.01 & MSP & Const & $\mathbf{.78 \spm .00}$ & \textcolor{ForestGreen}{$.14 \spm .01$} & \textcolor{ForestGreen}{$.24 \spm .02$} & $\underline{.0001 \spm .0000}$ & \textcolor{ForestGreen}{$\underline{1.267 \spm .016}$} & \textcolor{ForestGreen}{$.35 \spm .00$} & \textcolor{ForestGreen}{$\mathbf{.001 \spm .000}$} & \textcolor{ForestGreen}{$.119 \spm .003$} & \textcolor{ForestGreen}{$.047 \spm .002$} & \textcolor{ForestGreen}{$.016 \spm .002$} & \textcolor{ForestGreen}{$\underline{.000 \spm .001}$} \\
TS (ours) & 0.01 & MSP & Ada & $\mathbf{.78 \spm .00}$ & \textcolor{ForestGreen}{$.49 \spm .00$} & \textcolor{red}{$.65 \spm .01$} & $\underline{.0001 \spm .0000}$ & \textcolor{ForestGreen}{$2.040 \spm .015$} & \textcolor{ForestGreen}{$.61 \spm .00$} & \textcolor{ForestGreen}{$.054 \spm .000$} & \textcolor{ForestGreen}{$.186 \spm .001$} & \textcolor{ForestGreen}{$.094 \spm .001$} & \textcolor{ForestGreen}{$.046 \spm .001$} & \textcolor{ForestGreen}{$.008 \spm .001$} \\
TS (ours) & 0.05 & APS & Const & $\mathbf{.78 \spm .00}$ & \textcolor{ForestGreen}{$.28 \spm .01$} & \textcolor{ForestGreen}{$.42 \spm .01$} & $\underline{.0001 \spm .0000}$ & \textcolor{ForestGreen}{$1.472 \spm .019$} & \textcolor{ForestGreen}{$.42 \spm .00$} & \textcolor{ForestGreen}{$.010 \spm .001$} & \textcolor{ForestGreen}{$.151 \spm .001$} & \textcolor{ForestGreen}{$.072 \spm .002$} & \textcolor{ForestGreen}{$.035 \spm .001$} & \textcolor{ForestGreen}{$.007 \spm .001$} \\
TS (ours) & 0.05 & APS & Ada & $\mathbf{.78 \spm .00}$ & \textcolor{ForestGreen}{$.39 \spm .01$} & \textcolor{ForestGreen}{$.56 \spm .01$} & $\underline{.0001 \spm .0000}$ & \textcolor{ForestGreen}{$1.823 \spm .015$} & \textcolor{ForestGreen}{$.53 \spm .00$} & \textcolor{ForestGreen}{$.040 \spm .000$} & \textcolor{ForestGreen}{$.163 \spm .004$} & \textcolor{ForestGreen}{$.077 \spm .002$} & \textcolor{ForestGreen}{$.034 \spm .002$} & \textcolor{ForestGreen}{$.003 \spm .001$} \\
TS (ours) & 0.05 & MSP & Const & $\mathbf{.78 \spm .00}$ & \textcolor{ForestGreen}{$\underline{.08 \spm .01}$} & \textcolor{ForestGreen}{$\mathbf{.13 \spm .01}$} & $\underline{.0001 \spm .0000}$ & \textcolor{ForestGreen}{$\mathbf{1.244 \spm .018}$} & \textcolor{ForestGreen}{$\mathbf{.32 \spm .00}$} & \textcolor{ForestGreen}{$\underline{.001 \spm .000}$} & \textcolor{ForestGreen}{$.094 \spm .005$} & \textcolor{ForestGreen}{$\underline{.028 \spm .003}$} & \textcolor{ForestGreen}{$\underline{.001 \spm .003}$} & \textcolor{red}{$.013 \spm .002$} \\
TS (ours) & 0.05 & MSP & Ada & $\mathbf{.78 \spm .00}$ & \textcolor{ForestGreen}{$.45 \spm .01$} & \textcolor{red}{$.65 \spm .01$} & $\underline{.0001 \spm .0000}$ & \textcolor{ForestGreen}{$1.991 \spm .014$} & \textcolor{ForestGreen}{$.59 \spm .00$} & \textcolor{ForestGreen}{$.051 \spm .001$} & \textcolor{ForestGreen}{$.181 \spm .002$} & \textcolor{ForestGreen}{$.089 \spm .001$} & \textcolor{ForestGreen}{$.042 \spm .001$} & \textcolor{ForestGreen}{$.007 \spm .001$} \\
TS (ours) & 0.1 & APS & Const & $\mathbf{.78 \spm .00}$ & \textcolor{ForestGreen}{$.28 \spm .00$} & \textcolor{ForestGreen}{$.41 \spm .01$} & $\underline{.0001 \spm .0000}$ & \textcolor{ForestGreen}{$1.479 \spm .018$} & \textcolor{ForestGreen}{$.42 \spm .00$} & \textcolor{ForestGreen}{$.010 \spm .000$} & \textcolor{ForestGreen}{$.149 \spm .001$} & \textcolor{ForestGreen}{$.069 \spm .001$} & \textcolor{ForestGreen}{$.035 \spm .001$} & \textcolor{ForestGreen}{$.006 \spm .000$} \\
TS (ours) & 0.1 & MSP & Const & $\mathbf{.78 \spm .00}$ & \textcolor{ForestGreen}{$\mathbf{.07 \spm .00}$} & \textcolor{ForestGreen}{$\underline{.21 \spm .08}$} & $\underline{.0001 \spm .0000}$ & \textcolor{ForestGreen}{$1.379 \spm .036$} & \textcolor{ForestGreen}{$\underline{.33 \spm .00}$} & \textcolor{ForestGreen}{$.001 \spm .000$} & \textcolor{ForestGreen}{$\underline{.068 \spm .005}$} & \textcolor{ForestGreen}{$\mathbf{.002 \spm .005}$} & \textcolor{ForestGreen}{$.025 \spm .005$} & \textcolor{red}{$.035 \spm .003$} \\
TS (ours) & 0.1 & APS & Ada & $\mathbf{.78 \spm .00}$ & \textcolor{ForestGreen}{$.37 \spm .01$} & \textcolor{ForestGreen}{$.52 \spm .01$} & $\underline{.0001 \spm .0000}$ & \textcolor{ForestGreen}{$1.783 \spm .026$} & \textcolor{ForestGreen}{$.51 \spm .01$} & \textcolor{ForestGreen}{$.037 \spm .001$} & \textcolor{ForestGreen}{$.157 \spm .003$} & \textcolor{ForestGreen}{$.072 \spm .001$} & \textcolor{ForestGreen}{$.033 \spm .001$} & \textcolor{ForestGreen}{$.001 \spm .001$} \\

TS (ours) & 0.1 & MSP & Ada & $\mathbf{.78 \spm .00}$ & \textcolor{ForestGreen}{$.39 \spm .04$} & \textcolor{ForestGreen}{$.64 \spm .02$} & $\underline{.0001 \spm .0000}$ & \textcolor{ForestGreen}{$1.886 \spm .062$} & \textcolor{ForestGreen}{$.55 \spm .03$} & \textcolor{ForestGreen}{$.044 \spm .005$} & \textcolor{ForestGreen}{$.173 \spm .006$} & \textcolor{ForestGreen}{$.081 \spm .005$} & \textcolor{ForestGreen}{$.041 \spm .003$} & \textcolor{ForestGreen}{$.005 \spm .002$} \\
TS (ours) & 0.2 & APS & Const & $\mathbf{.78 \spm .00}$ & \textcolor{ForestGreen}{$.30 \spm .00$} & \textcolor{ForestGreen}{$.42 \spm .01$} & $\underline{.0001 \spm .0000}$ & \textcolor{ForestGreen}{$1.510 \spm .015$} & \textcolor{ForestGreen}{$.43 \spm .00$} & \textcolor{ForestGreen}{$.013 \spm .000$} & \textcolor{ForestGreen}{$.146 \spm .002$} & \textcolor{ForestGreen}{$.071 \spm .001$} & \textcolor{ForestGreen}{$.034 \spm .001$} & \textcolor{ForestGreen}{$.005 \spm .001$} \\
TS (ours) & 0.2 & MSP & Const & $\mathbf{.78 \spm .00}$ & \textcolor{ForestGreen}{$.10 \spm .00$} & \textcolor{ForestGreen}{$.30 \spm .02$} & $\underline{.0001 \spm .0000}$ & \textcolor{ForestGreen}{$1.436 \spm .023$} & \textcolor{ForestGreen}{$.35 \spm .00$} & \textcolor{ForestGreen}{$.009 \spm .000$} & \textcolor{ForestGreen}{$.090 \spm .002$} & \textcolor{ForestGreen}{$.035 \spm .003$} & \textcolor{ForestGreen}{$\mathbf{.001 \spm .002}$} & \textcolor{red}{$.019 \spm .002$} \\
TS (ours) & 0.2 & APS & Ada & $\mathbf{.78 \spm .00}$ & \textcolor{ForestGreen}{$.37 \spm .01$} & \textcolor{ForestGreen}{$.50 \spm .01$} & $\underline{.0001 \spm .0000}$ & \textcolor{ForestGreen}{$1.793 \spm .027$} & \textcolor{ForestGreen}{$.51 \spm .01$} & \textcolor{ForestGreen}{$.038 \spm .001$} & \textcolor{ForestGreen}{$.152 \spm .001$} & \textcolor{ForestGreen}{$.073 \spm .001$} & \textcolor{ForestGreen}{$.031 \spm .001$} & \textcolor{ForestGreen}{$\mathbf{.000 \spm .001}$} \\
TS (ours) & 0.2 & MSP & Ada & $\mathbf{.78 \spm .00}$ & \textcolor{ForestGreen}{$.37 \spm .01$} & \textcolor{ForestGreen}{$.62 \spm .01$} & $\underline{.0001 \spm .0000}$ & \textcolor{ForestGreen}{$1.813 \spm .017$} & \textcolor{ForestGreen}{$.52 \spm .00$} & \textcolor{ForestGreen}{$.040 \spm .001$} & \textcolor{ForestGreen}{$.167 \spm .002$} & \textcolor{ForestGreen}{$.086 \spm .001$} & \textcolor{ForestGreen}{$.041 \spm .000$} & \textcolor{ForestGreen}{$.005 \spm .000$} \\
\bottomrule
\end{tabular}
}
\end{table*}

\subsection{Takeaways across datasets}
  Based on the observed results, we can make the following observations:
  \begin{enumerate}
    \item Our methods most closely match the desired \(1-\alpha\) coverage across CIFAR-100, ImageNet, and iNaturalist.

    \item We consistently attain the best or near-best CMCE, highlighting the benefit of calibrating cumulative mass relative to conformal sets.

    \item  In large-\(K\) settings (ImageNet, iNaturalist), OvR binary methods often degrade calibration and even accuracy. In contrast, our methods remain stable and competitive on key calibration metrics while substantially improving set-mass alignment.
  \end{enumerate}

\end{document}